\documentclass[journal]{IEEEtranTIE}
\usepackage{graphicx}
\usepackage{picinpar}
\usepackage{amsmath}
\usepackage{amssymb}
\usepackage{amsfonts}
\usepackage{subcaption}
\usepackage{flushend}
\usepackage[latin1]{inputenc}
\usepackage{colortbl}
\usepackage{soul}
\usepackage{multirow}
\usepackage{multicol}
\usepackage{booktabs}
\usepackage{pifont}
\usepackage{color}
\usepackage{alltt}
\usepackage{dsfont}
\usepackage[hidelinks]{hyperref}
\usepackage{enumerate}
\usepackage{siunitx}
\usepackage{breakurl}
\usepackage{epstopdf}
\usepackage{pbox}

\usepackage{changes}
\usepackage{lipsum}
\definechangesauthor[name={Per cusse}, color=orange]{per}

\usepackage[ruled, linesnumbered]{algorithm2e}
\usepackage[hidelinks]{hyperref}

\begin{document}
\title{PSE-Match: A Viewpoint-free Place Recognition \\ Method with Parallel Semantic Embedding}

\author{
	\vskip 1em
	{Peng Yin$^1$, \textit{Graduate Student Member, IEEE},  Lingyun Xu$^{1*}$, Ziyue Feng$^3$, \\ Anton Egorov$^2$ and Bing Li$^3$, \textit{Member, IEEE}}

	\thanks{
		{
		Manuscript received July, 2020; revised February, 2021; accepted August, 2021.
		\newline \indent This work was supported in part by the U.S. Department of Transportation (grant 69A3551747117) through University Transportation Center - Center for Connected Multimodal Mobility (C2M2) at Clemson University.
		\newline \indent Peng Yin and Lingyun Xu are with the Robotics Institute, Carnegie Mellon University, Pittsburgh, PA 15213, USA.{(pyin2@andrew.cmu.edu, hitmaxtom@gmail.com)}.
		Anton Egorov is with the Skolkovo Institute of Science and Technology, Moscow, 121205, Russia. (Anton.Egorov@skoltech.ru).
        Ziyue Feng and Bing Li are with the Department of Automotive Engineering, Clemson University International Center for Automotive Research (CU-ICAR), Greenville, SC 29607 USA. (e-mails: zfeng, bli4@clemson.edu).
        \newline \indent * Lingyun Xu is the corresponding author (hitmaxtom@gmail.com).
		}
	}
}

\maketitle

\begin{abstract}
Accurate localization on the autonomous driving cars is essential for autonomy and driving safety, especially for complex urban streets and search-and-rescue subterranean environments where high-accurate GPS is not available.
However current odometry estimation may introduce the drifting problems in long-term navigation without robust global localization.
The main challenges involve scene divergence under the interference of dynamic environments and effective perception of observation and object layout variance from different viewpoints.
To tackle these challenges, we present PSE-Match, a viewpoint-free place recognition method based on parallel semantic analysis of isolated semantic attributes from 3D point-cloud models.
Compared with the original point cloud, the observed variance of semantic attributes is smaller.
PSE-Match incorporates a divergence place learning network to capture different semantic attributes parallelly through the spherical harmonics domain.
Using both existing benchmark datasets and two in-field collected datasets, our experiments show that the proposed method achieves above $\textbf{70\%}$ average recall with top one retrieval and above $\textbf{95\%}$ average recall with top ten retrieval cases. 
And PSE-Match has also demonstrated an obvious generalization ability with limited training dataset.

\end{abstract}



\definecolor{limegreen}{rgb}{0.2, 0.8, 0.2}
\definecolor{forestgreen}{rgb}{0.13, 0.55, 0.13}
\definecolor{greenhtml}{rgb}{0.0, 0.5, 0.0}

\section{Introduction}


    \IEEEPARstart{M}{apping} and localization are fundamental functionalities for autonomous driving, and have been well explored over the past decade~\cite{Survey:VSLAM}. 
    The autonomous driving system usually includes localization module~\cite{Auto:larson2019autonomous}, perception module~\cite{Auto:perception}, planning and control modules~\cite{Auto:control}.
    As one of the core module of simultaneously localization and mapping (SLAM), robust place recognition ability can increase the localization accuracy by reducing odometry drift via providing re-localization within a given prior map (a.k.a robot kidnapped problem).
    However, place recognition is still a challenging task due to the scene variation caused by dynamic environments, appearance and layout changes from different viewpoints.
    In the above extreme scenarios, robust re-localization is one of the main bottlenecks for autonomous driving. 
    
    \begin{figure}[t]
    	\centering
         \includegraphics[width=0.95\linewidth]{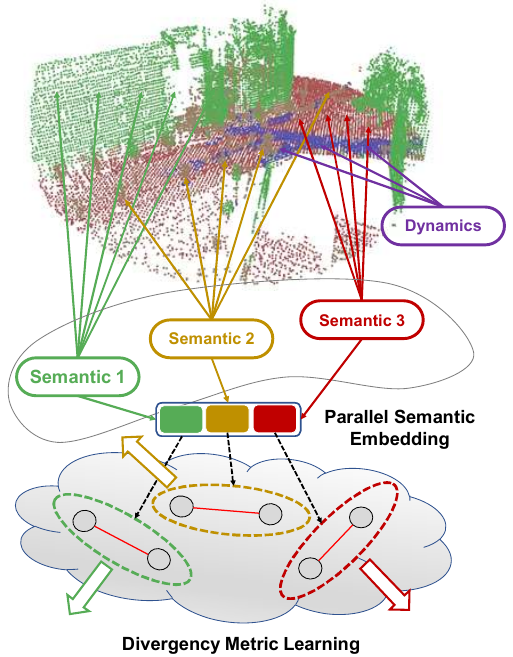}
    	\caption{3D place recognition under variant viewpoints and dynamic environments is a challenging problem.
    	PSE-match can achieve viewpoint-free place recognition by leveraging different static 3D semantic attributes through a parallel semantic embedding network.
    	We further introduce a divergence-metric learning module to explore various perspectives through different semantic embedding networks.}
    	\label{fig:idea}
    \end{figure}
    
    Visual localization and place recognition have been significantly explored in the computer vision community.
    Many existing researches adopt local keypoints and global appearance by overcoming the differences of perspective in place recognition.
    FAB-MAP~\cite{VPR:FABMAP} and ORB-SLAM~\cite{VPR:ORB-SLAM} utilize Bag-of-visual-Words (BoVW)~\cite{FeatureCapturer:BoW2} as place recognition module to encode each keyframe into an order-invariant vocabulary tree for local features~\cite{FEATURE:ORB,FEATURE:SURF}.
    Appearance-based methods, such as NetVLAD~\cite{PR:netvlad} and SeqSLAM~\cite{VPR:SeqSLAM}, compress single or continuous frame images into a global scene identifier to improve the robustness under perspective variations.
    However, there are vast differences between keypoint-based local features in perspective transformation. On the other hand, the state-of-the-art appearance-based methods can only deal with limited perspective differences with regarding to translation or orientation.
    Compared with visual camera features, 3D Light-Detection-and-Ranging (LiDAR) sensing data are more consistent for environmental changes and viewpoints differences.
    SegMap~\cite{segmap} mainly consider extract different 3D segments for place recognition, but the limitation for this method is that they need rich 3D geometry structures for segment extraction, which cannot be always satisfied.
    With the advanced 3D deep learning (e.g. PointNet~\cite{PointNet,pointnet++}), many recent researches~\cite{PR:pointnetvlad,PR:PCAN,PR:LPDNet} employ place descriptors from PointNet-based feature extraction and achieve top scene recognition performance results in open datasets.
    Since the original PointNetVLAD~\cite{PR:pointnetvlad} method and its variants are mainly designed to handle the interference of translation movement for scene recognition, the aforementioned point-based methods are often unable to overcome the interference from both translation and orientation changes at the same time.

    Therefore, we propose PSE-Match, a viewpoint free 3D scene recognition method. 
    Our proposed method can handle both locally translation-(XY-plane) and orientation-(heading) differences simultaneously through the spherical convolution.
    Besides, since the accuracy and stability of place recognition model are profoundly affected by robot's trajectory, we also design a dynamic octree mapping method to generate stable local map for inferencing.
    Our method utilizes a parallel semantic embedding network to encode different semantic attributes into independent place descriptors. 
    The detected semantic attributes enable robust features extraction, thus more reliable consistency for feature matching using only the static information. 
    In addition, it allows our feature extraction model to learn different semantic factors independently.
    As shown in Fig.~\ref{fig:idea}, we extract four semantic attributes for the point-cloud, including roads, buildings, static objects, and dynamic objects (e.g., cars and pedestrians).
    We only choose non-dynamic ones for feature extraction to avoid the disturbance from dynamic environments.
    Thus, the extracted feature descriptors can take advantage of semantic attributes by avoiding dynamic disturbance and improving semantic constraints.
    The main contributions of this paper include the following:
    \begin{itemize}
        \item A viewpoint-free 3D place recognition pipeline that simultaneously leverage the viewpoint-invariant place features from different semantic attributes.
    	\item A parallel semantic embedding network that encode semantic attributes into place descriptors through a spherical convolution network, which can extract orientation-invariant descriptors from spherical projections.
        \item A divergence learning metric to enhance the extracted place descriptors invariant to locally translation viewpoints difference, where individual attributes are more consistent than entire scenarios in translation variants. 
        \item Demonstration of the limitation of the state-of-the-art LiDAR-based place recognition approaches that are either focusing on orientation- or translation-invariance, while real large-scale place recognition tasks are with both changes.
	\end{itemize}
	
    We conducted a comprehensive recognition performance analysis on two public datasets, and two in-field collected datasets under orientation-translation difference. The proposed method achieves higher precision-recall than existing LiDAR-based place recognition methods.
    The remaining of the paper is structured as follows.
    In Section~\ref{sec:related_work}, we conduct a survey for 3D place recognition approaches and semantic-enhanced place recognition methods.
    Section~\ref{sec:method} introduces the overall of the proposed method, and elaborates the details of PSE-Match design.
    Experimental setup and performance analysis on different datasets are presented in Section~\ref{sec:experiments}, and quantitative analysis and comparison are conducted between PSE-Match and current state-of-the-art place recognition methods.
    Section~\ref{sec:conclusions} concludes the results and discusses the future works.

\section{Related Work}
\label{sec:related_work}

    A holistic review for 3D-based place recognition methods is presented in this section. For general 3D place recognition using geometric information, handcrafted and statistic-based features have been broadly explored, and learning-based features were also introduced to improve the feature robustness. More recently, 3D semantic-enhanced place recognition shows the potential of higher accuracy by enhancing the feature descriptor retrieving and matching.
    
    Unlike SIFT~\cite{lowe1999object} and SURF~\cite{bay2006surf} in image fields, hand-crafted features for the 3D point-cloud did not achieve such great success. 
    Keypoint based feature Spin-image~\cite{AG:SpinImageNew}, distance and angle based feature ESF~\cite{esf}, structure based Scan Context\cite{scan_context}, and histogram-based feature SHOT~\cite{shot} are both focused on local areas and not suitable for extracting global descriptors. 
    Fast Histogram~\cite{rohling2015fast} uses a histogram to represent the entire point-cloud distribution, M2DP~\cite{he2016m2dp} projects all points to several planes, they are designed to obtain a global descriptor, however, requires high density and complete data, and sensitive to noises.
    Recently, learning-based point-cloud descriptors have shown outstanding performance and gradually replacing handcraft features. With the breakthrough of the convolutional neural networks, NetVLAD~\cite{PR:netvlad} provides a successful learning-based paradigm to aggregate local features to form a global descriptor. 
    Since the NetVLAD~\cite{PR:netvlad} only accepts 2D image inputs, to leverage the illumination invariant advantage of the LiDAR sensor, PointNetVLAD~\cite{PR:pointnetvlad} incorporates the iconic point-cloud processing network PointNet~\cite{PointNet} to achieve an end to end learning-based point-cloud global feature extractor. 
    To further improve the performance of the PointNetVLAD~\cite{PR:pointnetvlad}, PCAN\cite{PR:PCAN} adds a context-aware attention module to give different weights to different contributing contexts, LPD-Net\cite{PR:LPDNet} uses the PointNet++\cite{pointnet++} to introduce more multi-scale feature extracting ability.
    
    These PointNet~\cite{PointNet} and PointNet++~\cite{pointnet++} based methods are translation-invariant, but sensitive to orientation changes. LocNet~\cite{locnet} uses a handcraft rotation-invariant intermediate representation as input, but it lost too much information from the original point-cloud. 
    SphereCNN~\cite{spherecnn} introduced a spherical convolution to efficiently obtain orientation-equivariant features. Inspired by it, we project the point-cloud with semantic information to a sphere space to get the orientation-equivariant features from spherical harmonics, then use a feature clustering operation to convert it to orientation-invariant global descriptors.
    
    Most of the existing place recognition methods are based on pure geometric or appearance information, which can't provide conceptual knowledge to support high-level complex tasks, and sensitive to illumination, season, and view point changes. 
    With the vigorous development and satisfactory performance of the deep learning based 2D~\cite{maskrcnn} and 3D~\cite{mink} semantic segmentation technologies, incorporating semantic information to achieve robust and accurate place recognition becomes possible. 
    Semantic information is much more robust than appearance information. 
    Johannes's work~\cite{schonberger2018semantic} proves that semantic features from 2D images can achieve successful localization under extreme appearance changes, which is challenging for appearance-based features.
    There are many works use semantic information to enhance feature descriptors. 
    Garg's work~\cite{semantic-geo} concatenates the appearance-based descriptor and the semantics-based descriptor, then normalize it to get the final global descriptor.

    In our work, we extract the 3D segmentation via SqueezeSeg~\cite{3DSegL:wu2018squeezeseg} and project the semantic segmentation result along with geometry to the sphere space to extract orientation-invariant global feature descriptor, which enhances place features robustness by focusing on different types of semantics.

\section{The Proposed Method}
\label{sec:method}

    \subsection{Overview}
\begin{figure}[ht]
        \centering
        \includegraphics[width=0.85\linewidth]{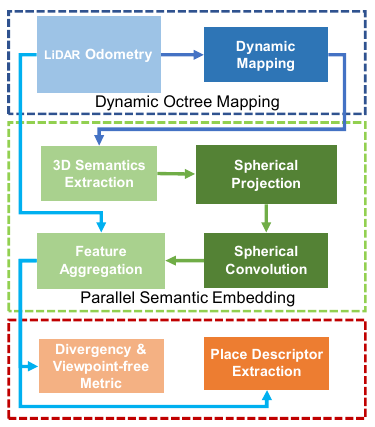}
        \caption{The framework pipeline of PSE-Match.
        The framework mainly includes three modules: 
        a dynamic octree-based mapping method for dense map generation (dashed-blue block),
        a parallel semantic embedding module for place descriptor extraction (dashed-green block), 
        and a divergence metric-based viewpoint-free feature learning module 
        to incorporate different aspects learned from the parallel semantic embedding (dashed-red block).}
        \label{fig:pipeline}
    \end{figure}

    As depicted in Fig.~\ref{fig:pipeline}, the framework of PSE-Match is mainly composed of three modules: Dynamic Octree mapping (DOM), Parallel Semantic Embedding (PSE), and Divergence enhanced Feature Learning (DFL).
    Because of the sparsity of a single LiDAR scan, we implement a dynamic octree mapping mechanism to generate dense local maps.
    By applying a modified LiDAR Odometry method~\cite{LOAM:Lego-LOAM} to estimate the transformation information for sequences of LiDAR inputs, and we generate a dense local map for each place by accumulating LiDAR scans via an octree mapping procedure~\cite{LPR:OCTOMAP}.
    In Parallel Semantic Embedding networks, we collect semantic attributes from a pre-trained 3D segmentation network~\cite{3DSegL:wu2018squeezeseg} and parallelly extract semantic-independent descriptors by designing a spherical convolution network.

    Finally, the learned descriptors are further enhanced by a divergence learning metric, which can explore the different semantic attributes in viewpoint-free place feature learning.

\subsection{Dynamic Octree Mapping}

\label{sec:dom}
The single scan of LiDAR input is too sparse to extract valid appearance feature, here we utilize a dynamic octree data structure to incrementally generate the dense local map from continuous frame scans. 
An octree is a tree type data structure, where the sub-children of a parent node divide the space into eight equal ones until the size of a leaf node reaches a given granularity.
DOM models the local map around the robot coordinate by fixing the map scale and updating the map based on the odometry estimation. 
Odometry error between maps $M_{t-1}$ and $M_{t}$ may introduce additional uncertainty for each leaf node's occupancy, which leads to mapping blurring problem.
This problem can be alleviated using our dynamic octree mapping mechanism by taking into account of odometry uncertainty for the map updating.
We first describe the map measurement updating step in static octree updating by accumulating LiDAR inputs, 
then introduce dynamic octree updating method with the additional motion updating step. 

\subsubsection{Measurement Updating}
\label{sec:bom}
In traditional octree mapping method~\cite{LPR:OCTOMAP}, the occupancy beliefs of leaf nodes are updated by the log-odds method described in~\cite{ROBOT:probrobo}. 
For the leaf node $n$ with given sequential measurements $z_{1:t}$, the occupancy estimation $P(n|z_{1:t})$ can be calculated by
\begin{align}
&P(n|z_{1:t}) =  \nonumber \\ 
&[1+\frac{1-P(n|z_{t})}{P(n|z_{t})}\cdot \frac{1-P(n|z_{1:t-1})}{P(n|z_{1:t-1})}\cdot \frac{P(n)}{1-P(n)}]^{-1} \label{eq:measurement_update}
\end{align}
where $P(n)$ is the prior occupancy probability for the leaf node $n$, and $P(n|z_{t})$ is the occupancy belief based on the current observation. 
The log-odds function $L(x)$ over a probability distribution $P(x)$ is defined by
\begin{align}
L(x)=\log[\frac{P(x)}{1-P(x)}] \label{eq:logodds}
\end{align}
Without loss of generality, we can assume the initial occupancy $P(n)$ is $0.5$, we can combine Eq.~\ref{eq:measurement_update} and Eq.~\ref{eq:logodds} into \begin{align}
&L(n|z_{1:t}) = L(n|z_{1:t-1}) + L(n|z_{t})
\label{eq:simple_mup}
\end{align}
Then, the original occupancy updating problem is now transformed into a linear accumulation operation. 
This mechanism guides us in map center updating, where we need to modify the occupancy of each voxel based on the odometry estimation. 

    \begin{figure}[t]
    \centering
    \includegraphics[width=\linewidth]{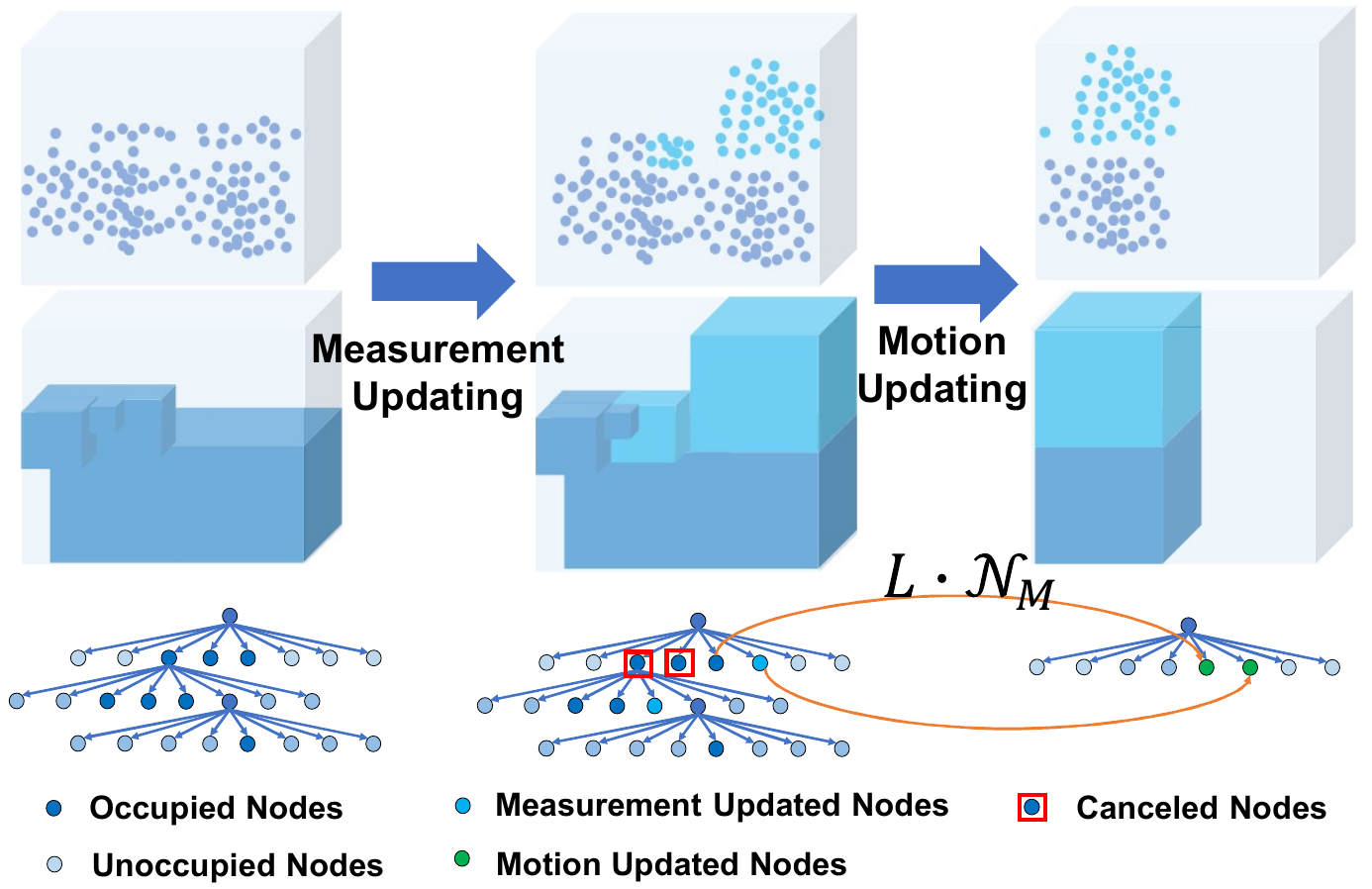}
    \caption{
            Map updating in the Dynamic Octree Mapping (DOM).
            Except the measurement updating step in traditional octree mapping method, DOM contains an additional motion updating step.
    		The first row shows 3D voxel changes during these two steps, the second row shows inner changes of octree structure, and the third row shows the changes of occupancy probability of occupied leaf nodes under different motion error and measurement error.
    		In the additional motion updating step, the occupancy of octree maps are further modified according to the covariance of estimated robot position according to Eq.~\ref{eq:motion}.
        }
    \label{fig:dom}
    \end{figure}

    \begin{figure*}[!th]
    	\centering
    	\includegraphics[width=\linewidth]{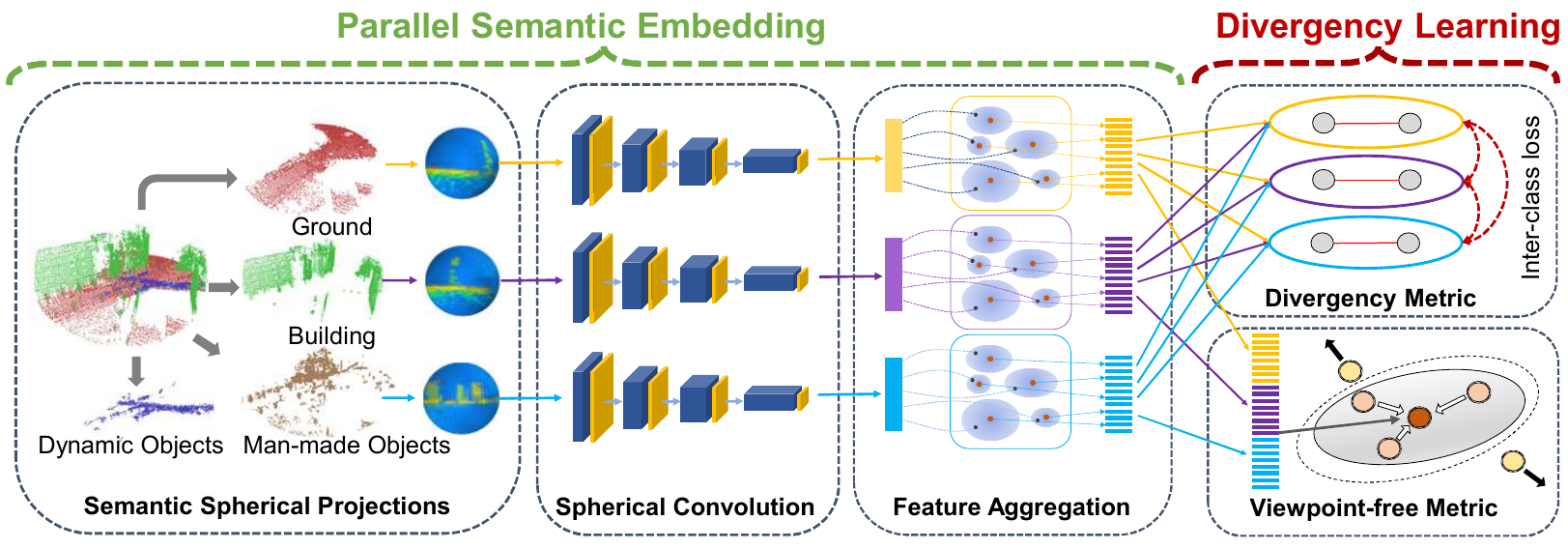}
    	\caption{The network structure of PSE-Match.
        The training procedure of method includes the parallel semantic embedding step and a divergence learning step. 
        In the parallel semantic embedding step, we only embed the static objects (i.e., ground, building and man-made structures) for orientation-invariant place feature learning.
        In the divergence learning step, we introduce two learning metrics: a viewpoint-free metric learning which shorten distances of features with only orientation- or locally translation- differences, and a regularization loss metric to capture different semantic aspects of original 3D point-cloud.
        }
    	\label{fig:pseNet}
    \end{figure*}

\subsubsection{Motion Updating}
\label{sec:dmu}
We cope with the motion updating error by modifying the log-odds updating in the occupancy updating step in Eq.~\ref{eq:simple_mup}.
In the motion updating step, the occupancy of each leaf node from map $M_{t-t_{p}}$ to map $M_{t}$, is updated with a motion error model $\mathcal{N}_{M}(0, \sigma_{t_{p}})$.
The covariance $\sigma_{t_{p}}$ can be evaluated based on the LiDAR odometry estimation results~\cite{LOAM:zhang2014loam}, and the more aggressive the motion, the larger the covariance value. 
A high confidence map can be obtained when vehicles travel in flat areas smoothly and vice versa.
Then we can update the original log-odds ${L}(n|z_{1:t})$ by incorporating additional motion error
\begin{align}
\hat{L}(n|z_{1:t}) = L(n|z_{1:t})\cdot \mathcal{N}_{M}(0, \sigma_{t_{p}}) \label{eq:motion}
\end{align}

Referring to the new log-odds $\hat{L}(n|z_{1:t})$, the new occupancy $\hat{P}(n|z_{1:t})$ can be updated with Eq.~\ref{eq:motion},
\begin{align}
\hat{P}(n|z_{1:t}) = \frac{\exp(\hat{L}(n|z_{1:t}))}{\exp(\hat{L}(n|z_{1:t}))+1}\label{newleafnode}
\end{align}

Since $\|\hat{L}(n|z_{1:t})\| \leq \|L(n|z_{1:t})\|$, by comparing the original $P(n|z_{1:t})$ with updated probability $\hat{P}(n|z_{1:t})$, we have,
\begin{align}
P(n|z_{1:t}) > 0.5 \Rightarrow P(n|z_{1:t}) > \hat{P}(n|z_{1:t}) > 0.5  \\
P(n|z_{1:t}) < 0.5 \Rightarrow P(n|z_{1:t}) < \hat{P}(n|z_{1:t}) < 0.5 \nonumber \label{compareToOldLeafnode}
\end{align}
Therefore, based on Eq.~\ref{newleafnode} and Eq.~\ref{compareToOldLeafnode}, the new updated probability $\hat{P}(n|z_{1:t})$ reduces the occupancy beliefs for both occupied and unoccupied leaf nodes, i.e., the motion error model $\mathcal{N}_{M}(0, \sigma_{t_{p}})$ introduce uncertainty in the motion updating step. Therefore, in the case of low odometry confidence, the map updating can be alleviated through this step.

\subsection{Parallel Semantic Embedding}
\label{sec:PSE}
In this section, we will describe the three modules for parallel semantic embedding: semantic spherical projection, spherical feature extraction and place feature aggregation.
The conventional spherical projection for all the point-clouds will loss geometry information of the original 3D point-cloud data~\cite{spherecnn}. In stead, we propose a novel approach of constructing a multi-layer spherical projections from different semantic attributes. It allows independent semantic attribute feature extraction, while maintaining the synergy of all attributes in the divergency learning as shown in section \ref{sec:DMFL}.

\BlankLine
\subsubsection{Semantic Spherical Projection}
\label{sec:SSP}

    In our PSE-Match, we project the 3D point-cloud on spherical perspectives for viewpoint-free feature extraction.
    The attributes are extracted by a pre-trained 3D semantic segmentation network~\cite{3DSegL:wu2018squeezeseg}, and have high consistence to environmental changes.
    As we can see the first module in Fig.~\ref{fig:pseNet}, we extract semantic elements, e.g., ground, structures and man-made objects as static objects;
    on the other hand, dynamic objects, such as cars and pedestrians etc, are ignored at the early stage of PSE-Match. 
    We generate a two-channel spherical representation for each static semantic attribute.
    We define spherical image $S^2$ as the set of points $x \in \mathbb{R}^3$ in the unit sphere. 
    We can parameters $S^2$ by spherical coordinates $\theta \in [0, 2\pi]$ and $\phi \in [0, \pi]$.
    Given a resolution $n$, we divide the flattened sphere plane into $n\times n$ equiangular spherical grids indexed by $(\theta_i, \phi_j)$.
    Here $\theta_i = 2\pi{i}/n, \phi_j = \pi{j} / n$, $0\leq{i, j}\leq{n-1}$.
    We take $d_{ij}$ which is the distance from center to the closest point in spherical grid $(\theta_i, \phi_j)$ as the first channel. 
    Using sine value of the angle $\alpha_{i, j}$ between ray towards the closest point and normal direction in spherical grid $(\theta_i, \phi_j)$, we get the second channel.
    Formally, the two-channel spherical representation $z_{A_{l}}$ of the semantic attribute $A_{l}$ can be written as: $z_{A_{l}}(\theta_i, \phi_j) = [d_{ij}, \sin{\alpha_{ij}}], 0\leq{i, j}\leq{n-1}$.

    \begin{figure*}[!th]
        \centering
        \includegraphics[width=\linewidth]{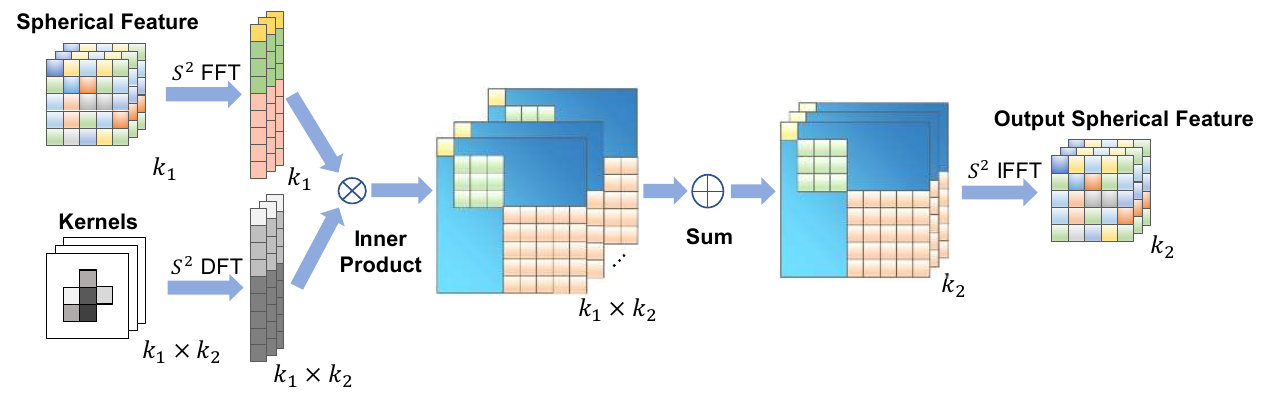}
        \caption{The spherical convolution module.
        With the given spherical feature $f$ and a kernel signal $h$, we first transform them into the harmonic domain ($H_f$, $H_h$)with the Fast Fourier transform (FFT) and Discrete Fourier Transform (DFT) respectively.
        In spherical convolution, the convolution operation is equal to the point-wise product of harmonic coefficients, and the shperical feature outputs can be formulated by Inverse Fast Fourier Transform (IFFT) operation.}
        \label{fig:sphnet}
    \end{figure*}

    \BlankLine
    \subsubsection{Spherical Feature Extraction}
    \label{sec:SFE}
    Representing data in the spherical domain can be quite natural and effective in analyzing 3D point-cloud; however, it is difficult to adopt traditional planner convolution network in such manifolds, because the spaces between adjacent points in the spherical domain are not uniform.
    Instead of planner convolution, we apply the spherical convolution on the spherical representations for orientation-invariant feature extraction.
    Spherical convolution avoids space-varying distortions in the Euclidean space by convolving spherical signals in the harmonic domain.
    The mathematical model of spherical convolution into the harmonic domain shows its orientation-equivalent.
    Here we firstly define the required mathematical concepts in the spherical convolution.
    We model spherical images and filters as functions: 
    $f:S^2 \rightarrow \mathbb{R}^K$,
    where $S^2$ is defined as the set of points $x \in \mathbb{R}^3$ in the unit sphere, and $K$ is the number of channels. 
    The set of rotations in $\mathbb{R}^3$ is called $SO(3)$, the "special orthogonal group".
    Each rotation can be represented by a $3 \times 3$ matrix that preserves distance (i.e. $\left\Vert R \circ x\right\Vert = \left\Vert x \right\Vert, x \in \mathbb{R}^3$), where $\circ$ denotes
    acting rotation $R$ on $x$.
    Based on the work in~\cite{Sphere:SO3_invariant}, we can introduce the rotation operator $L_R$ which acts on spherical signals (i.e. functions) $f: S^2 \rightarrow \mathbb{R}^K$ for $R \in SO(3)$ as:
    $[L_{R} \circ f](x) = f(R^{-1} \circ x)$.
    Thus the spherical convolution of signals $f$ and $h$ in the rotation group $SO(3)$ are defined as:
    \begin{align}
        [f \star_{SO(3)} h](\mathbf{R}) = & \int_{SO(3)}f(\mathbf{R}^{-1}\mathbf{Q})h(\mathbf{Q})d\mathbf{Q}.
    \end{align}
    where $\mathbf{R, Q} \in SO(3)$. 
    As the proof in ~\cite{cohen2018spherical}, spherical convolution is shown to be orientation-equivariant:
    \begin{align}
        [f \star_{SO(3)} [L_{Q}h](\mathbf{R})
        = & [L_{\mathbf{Q}}[f \star_{SO(3)} h]](\mathbf{R}).           
    \end{align}
    where $L_{\mathbf{Q}} (\mathbf{Q} \in SO(3))$ is a rotation operator for spherical signals.
    As depicted in Fig.~\ref{fig:sphnet}, the convolution of two spherical signals $f$ and $h$ in the harmonica domain are computed by three steps.
    We first expand $f$ and $h$ to their spherical harmonic basis $H_f$ and $H_h$, then compute the point-wise product of harmonic coefficients, and finally invert the spherical harmonic expansion.
    For more details, we suggest the reader refers to the original work in~\cite{Sphere:SO3_invariant}.

    \BlankLine
    \subsubsection{Place Feature Aggregation}
    \label{sec:VPDE}
        We leverage a feature clustering operation to convert the output of spherical convolution into an orientation-invariant place descriptor.
        Intuitively, there exists spatial similarity in output local descriptors of spherical convolution.
        Therefore, we cluster the local descriptors and take a sum of residuals (difference vector between descriptor and its corresponding cluster center) as a global place descriptor.
        Formally, given $N$ $D$-dimensional local descriptors $\{\mathbf{x}_i\}$ as input, and $K$ cluster centers (``visual words'') ${\mathbf{c}_k}$ as VLAD parameters, the $(j, k)$ element of output global descriptor $\mathbf{V}$:
        \begin{align}
            \mathbf{V}(j, k) = \sum_{i=1}^{N}\overline{a}_k(\mathbf{x}_i)(x_i(j) - c_k(j)),
        \end{align}
        where $x_i(j)$ and $c_k(j)$ are the $j$-th dimension of the $i$-th descriptor and $k$-th cluster center, respectively. 
        Here $\overline{a}_k(\mathbf{x}_i)$ has the following definition:
        \begin{align}
            \overline{a}_k(\mathbf{x}_i) = \frac{e^{\mathbf{w}_k^T}\mathbf{x}_i + b_k}{\sum_{k}e^{\mathbf{w}_{k}^T\mathbf{x}_i + b_{k}}},
        \end{align}
        where vector $\mathbf{w}_k$ and scalar $b_k$ are two learnable parameters.
        The clustered place features satisfy the permutation invariance of input local descriptors (order of $\{\mathbf{x}_i\}$), which makes the place descriptors invariant to orientation difference and locally translation difference.
        For the detailed analysis of VLAD please refer to the original paper~\cite{PR:vlad}

    \subsection{Divergence Metric-based Feature Learning}
    \label{sec:DMFL}
    To incorporate different aspects learned from the parallel semantic embedding network, we introduce a viewpoint-free divergence loss metric for place feature learning.
    This divergence loss includes two loss metric parts: a ``Viewpoint-free'' loss to enhance feature robustness to orientation- and local translation- variants, and a regularizing term to force the network to learn from different semantic attributes.
    
    For convenience of illustrating the loss functions, some necessary definitions are first described in following.
    Each training tuple in our training datasets consists of four components: $\mathcal{S} = [S_a, \{S_{rot}\}, \{S_{pos}\}, \{S_{neg}\}]$, where $S_a$ is the spherical representation of a local 3D scan,
    $\{S_{rot}\}$ is a set of spherical representations of 3D scans rotated with different angles ($[\ang{0},\ang{30},...\ang{330}]$).
    $\{S_{pos}\}$ denotes a set of spherical representations of 3D scans (``positive'') which are structurally similar to $\{S_a\}$,
    and $\{S_{neg}\}$ denotes a set of 3D scans (``negative'')
    which are structurally different to $\{S_a\}$.
    Ideally, the model can minimize two distances: $\delta_{pos}^j=d(f(S_a), f(S_{pos}^j))$ and $\delta_{pos_i}^{rot_j} = d(f(S_{rot}^j), f(S_{pos}^i))$, while maximizing: $\delta_{neg}^{j} = d(f(S_a), f(S_{neg}^j))$ and $\delta_{neg_i}^{rot_j} = d(f(S_{rot}^j), f(S_{neg}^i))$. 
    Here $f(.)$ is the function that encodes spherical representations into a global descriptors by PSE-Match, and $d(\cdot)$ denoted the squared Euclidean distance.
    
    \begin{figure}[t]
    	\centering
    	\includegraphics[width=\linewidth]{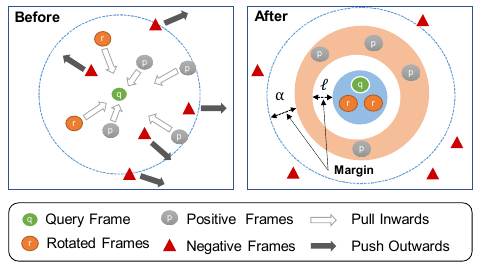}
    	\caption{The Viewpoint-free loss metric.
    	We provide the current query frame, rotated frames (same places with different orientations), positive frames (places with local translation differences) and negative frames (places with translation difference beyond certain distance) for viewpoint-free feature learning.
    	The ``Viewpoint-free'' metric can pull or push features of rotated, positive and negative frames towards the query frame based on the margin settings.
    	}
    	\label{fig:trip_loss}
    \end{figure}
    
    The ``Viewpoint-free'' loss metric contains a ``LazyRot'' loss for orientation-invariant feature learning and a ``LazyTrip'' loss for the locally translation-invariant.
    ``LazyRot'' loss is designed to maximize the feature distance between $f(S_{rot})$ and $f(S_{neg})$, while reduce feature difference between $f(S_{rot})$ and $f(S_{pos})$.
    Formally, the ``Lazy Rotation`` loss is denoted as:
    \begin{align}
        L_{LazyRot}(\mathcal{T}) = \max_{i,j,k}([\ell + \delta_{pos_i}^{rot_j} - \delta_{neg_k}^{rot_j}]_+),
    \end{align}
    where $[.]_+$ denotes the hinge loss and $\ell$ is a constant parameter to mark corresponding margin.
    With this soft loss constraints, PSE-Match can further reduce the feature similarity within local boundaries even under variant orientation differences.
    We apply a ``Lazy Triplet'' loss to maximize the feature distance between the local scan feature $f(S_{rot})$ and its corresponding ``negative'' features $f(S_{neg})$, which is represented as:
    \begin{align}
        L_{LazyTrip}(\mathcal{T}) = \max_j([\alpha + \delta_{pos} - \delta_{neg_j}]_+),
    \end{align}
    here $\alpha$ is constant threshold to control the margin.
    We employ the simultaneous supervision of ``Lazy Rotation" loss and ``Lazy Triplet" loss to learn the ``viewpoint-free" descriptors for the semantic attribute $A_{l}$ by:
    \begin{align}
        L_{free}^{A_{l}} = L_{LazyRot}^{A_{l}} + L_{LazyTrip}^{A_{l}}.
    \end{align}
    
    The final divergence loss metric is based on the each semantic attribute loss $L_{free}^{A_{l}}$ and an additional regularization term,
    \begin{align}
        L_{Divergence} = \sum_{l} L_{free}^{A_{l}} + \lambda \sum_{a,b}[m-D_{ia,ib}^2]_{+},
    \end{align}
    where $D_{ia,ib}$ represents a distance measurement between place descriptors learned by different semantic embedding branches, $m$ is the margin between two different leaners $a$ and $b$, and $\lambda$ is the weighting parameter to control the regularization strength.

    \begin{figure}[t]
    	\centering
    	\includegraphics[width=1.0\linewidth]{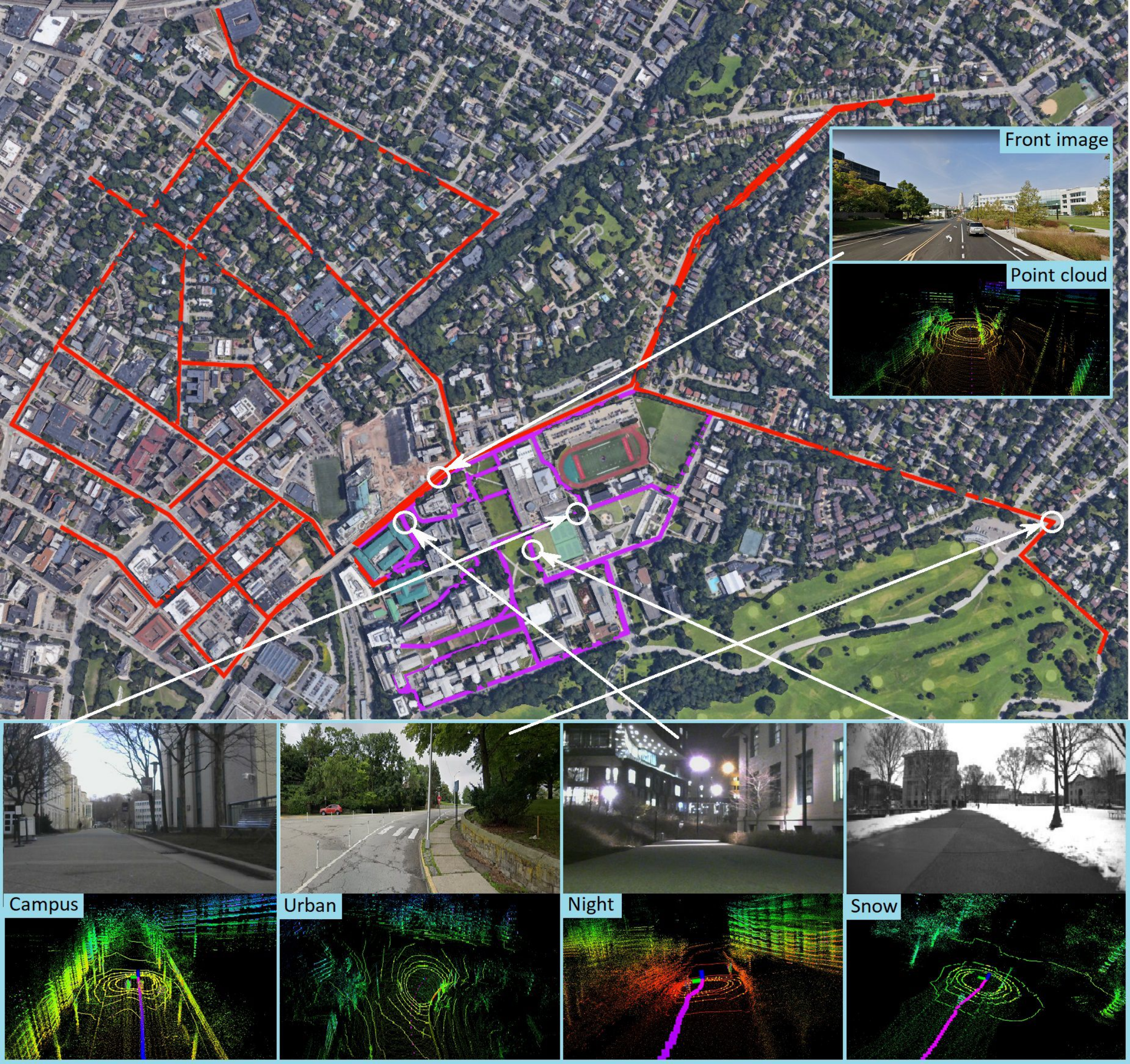}
    	\caption{The in-field datasets which contains a campus area and city-scale area.
    	All the datasets are collected by a data-recording platform with a Velodyne-VLP 16 LiDAR device, a front view camera, an IMU device, and a Global Navigation Satellite Systems (GNSS) system for ground truth positions.
    	Campus datasets includes $11$ trajectories around the campus of Carnage Mellon University (CMU).
    	The city-scale datasets contains $12$ trajectories outside the CMU campus in the city of Pittsburgh.
    	The data are recorded incrementally on different parts, different environment conditions and different tracks.}
    	\label{fig:map}
    \end{figure}

    \begin{figure*}[t]
        \centering
        \includegraphics[width=\linewidth]{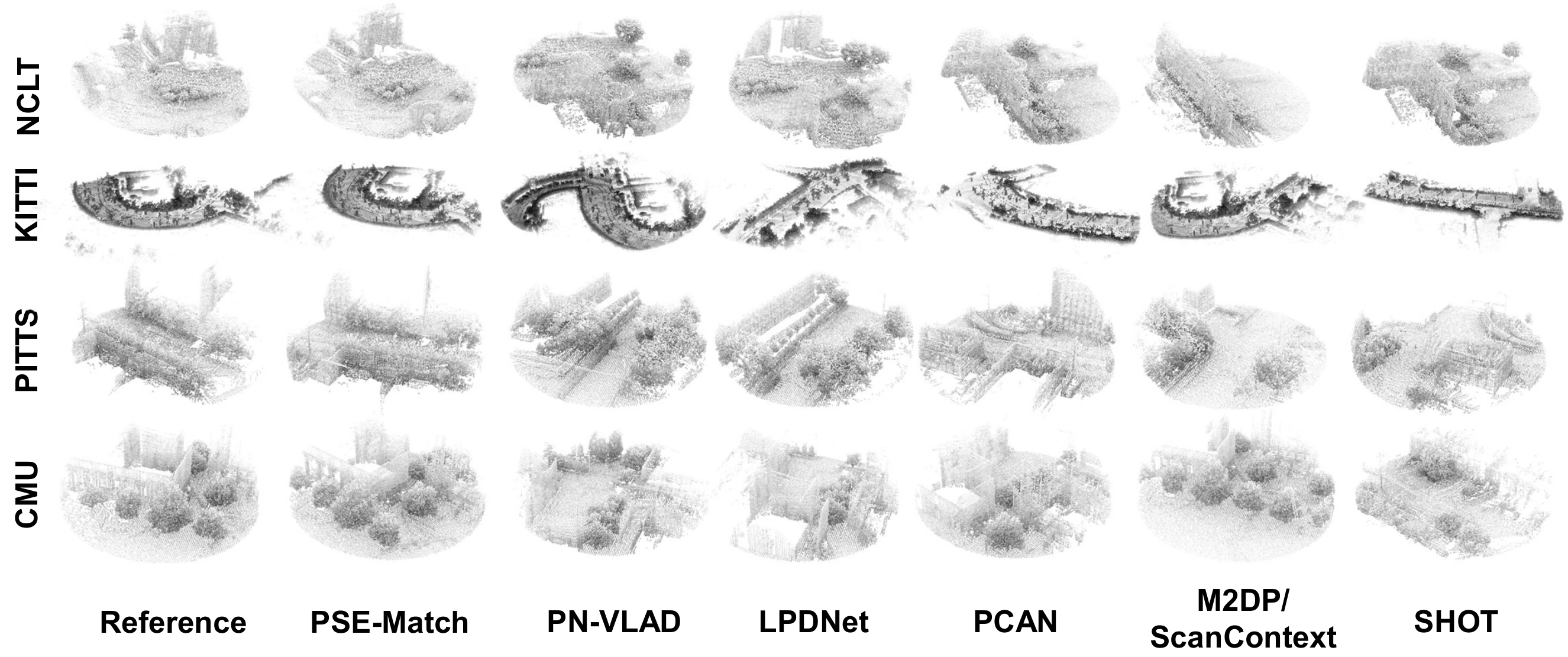}
        \caption{Place retrievals on two public available datasets and two in-field collected datasets.
        The first column shows the original reference maps on different datasets, and the following columns present the top one matched maps retrieved on testing frames by different methods.
        Both ScanContext and M2DP achieve similiar performances.}
        \label{fig:dataset}
    \end{figure*}

\section{Experiments}
\label{sec:experiments}
    This section compares the proposed method with both non-learning/learning-based 3D place recognition methods on two public datasets, and two in-field collected datasets.
    To generate our datasets, we designed a data-collecting platform, which contains a LiDAR device (Velodyne-VLP 16), an inertial measurement unit (Xsense MTi $30$, $0.5^{\circ}$ error in roll/pitch, $1^{\circ}$ error in yaw, $550m$W), a Global Navigation Satellite Systems (GNSS) ($10$Hz outputs), a mini PC (Intel NUC i7, $3.5$ GHz, $28$W) and an embedded GPU device (Nvidia Xavier, $8$G memory).
    We trained our network model on a GPU server with a Ubuntu 18.04 system with a single Nvidia 1080Ti GPU and $64$G RAM.
    In the rest of this section, we first introduce the datasets, the comparison of non-learning/learning-based methods, and evaluate manners and metrics.
    Then, we demonstrate the quantitative and qualitative comparison analysis of different methods under variant viewpoint configurations on different datasets.
    Finally, both resource requirements and computation efficiency are further evaluated for each methods.

    \bigskip
    \noindent \textbf{Datasets.} 
    The training and evaluation data includes two public available datasets and two in-filed collected datasets:
    \begin{itemize}
        \item \textbf{KITTI}~\cite{KITTI:Geiger} odometry datasets, which consists of 21 trajectories generated with Velodyne-64 LiDAR scanner, around the mid-size city of Karlsruhe. 
        We use trajectory $\{1\sim 8\}$ for network training, and $\{9,10\}$ for evaluation.
        \item \textbf{NCLT}~\cite{DATASET:NCTL} datasets, which consists of 103 trajectories generated with Velodyne-32 LiDAR scanner, around the mid-size city of Karlsruhe.
        We use trajectory $\{1\sim 8\}$ for network training, and $\{9,10\}$ for evaluation.
        \item \textbf{CMU dataset}, we created a \textit{Campus} dataset with 11 trajectories by driving our data capturing mobile platform to traverse a $2km$ outdoor route around the campus of our university. 
        We use trajectories $\{1\sim 9\}$ for network training, and $\{10,11\}$ for evaluation.
        \item \textbf{Pittsburgh dataset}, we created a City-scale dataset by mounting our data capturing system on the top of a car and incrementally traversing $12km$ trajectories in the city of Pittsburgh.
        We collected 12 trajectories, and we use trajectories $\{1\sim 10\}$ for network training, and $\{11,12\}$ for evaluation.
    \end{itemize}
    \begin{table}[ht]
        \centering
        \caption{Dataset frames splitting in training/evaluation.}
        \begin{tabular}{c c c c c}
        \toprule
         & \textit{KITTI} & \textit{NCLT} & \textit{CMU} & \textit{CITY} \\ \midrule
        Train       & $13,070$ & $15,971$ & $13,830$ & $10,972$ \\ 
        Evaluation  & $3,268$  & $3,993$  & $3,458$  & $2,744$  \\ 
        \bottomrule
        \label{table:dataset}
        \end{tabular}
    \end{table}
    The relationship between the two in-field dataset are shown in Fig.~\ref{fig:map}.
    Table.~\ref{table:dataset} shows the data splitting in the training and evaluation procedure on the four different datasets.
    We generate the reference and testing sequence from the same trajectory in the evaluation step for each dataset under different orientation and translation differences.
    And to evaluate the generalization ability of different learning-based methods, we introduce a \textit{refine} mode to train the network:
    we select every first two trajectories within four datasets for network training, and use the same testing data in Table~\ref{table:dataset} for evaluation.
    
    \bigskip
    \noindent \textbf{Evaluation  Methods.}
    To evaluate our approach's performance, we firstly compared it with the state-of-the-art learning-based place recognition approaches:  PointNetVLAD (PN-VLAD)\footnote{https://github.com/mikacuy/pointnetvlad}\cite{PR:pointnetvlad}, LPDNet\footnote{https://github.com/Suoivy/LPD-net}\cite{PR:LPDNet} and PCAN\footnote{https://github.com/XLechter/PCAN}\cite{PR:PCAN}.
    For non-learning based approaches, we also compare the frame-level descriptors ScanContext~\cite{scan_context}\footnote{https://github.com/irapkaist/scancontext}, M2DP~\cite{he2016m2dp}\footnote{https://github.com/LiHeUA/M2DP}, and 3D keypoint-based approaches SHOT and Fast Histogram from PCL~\cite{shot}\footnote{https://github.com/PointCloudLibrary/pcl} libaray.
    For all the above approaches, we provide the dense local maps generated by our dynamic octree mapping step, as shown in Section~\ref{sec:dom}, and extract the global place descriptors based on the different procedures of the above approaches.
    For the fair comparison, the inputs for all the above methods are processed with out dynamic octree mapping module. 
    And we also analysis the place recognition of our PSE-Match with and without the dynamic octree mapping module to investigate their differences.
    
    \begin{figure*}[t]
        \centering
        \includegraphics[width=\linewidth]{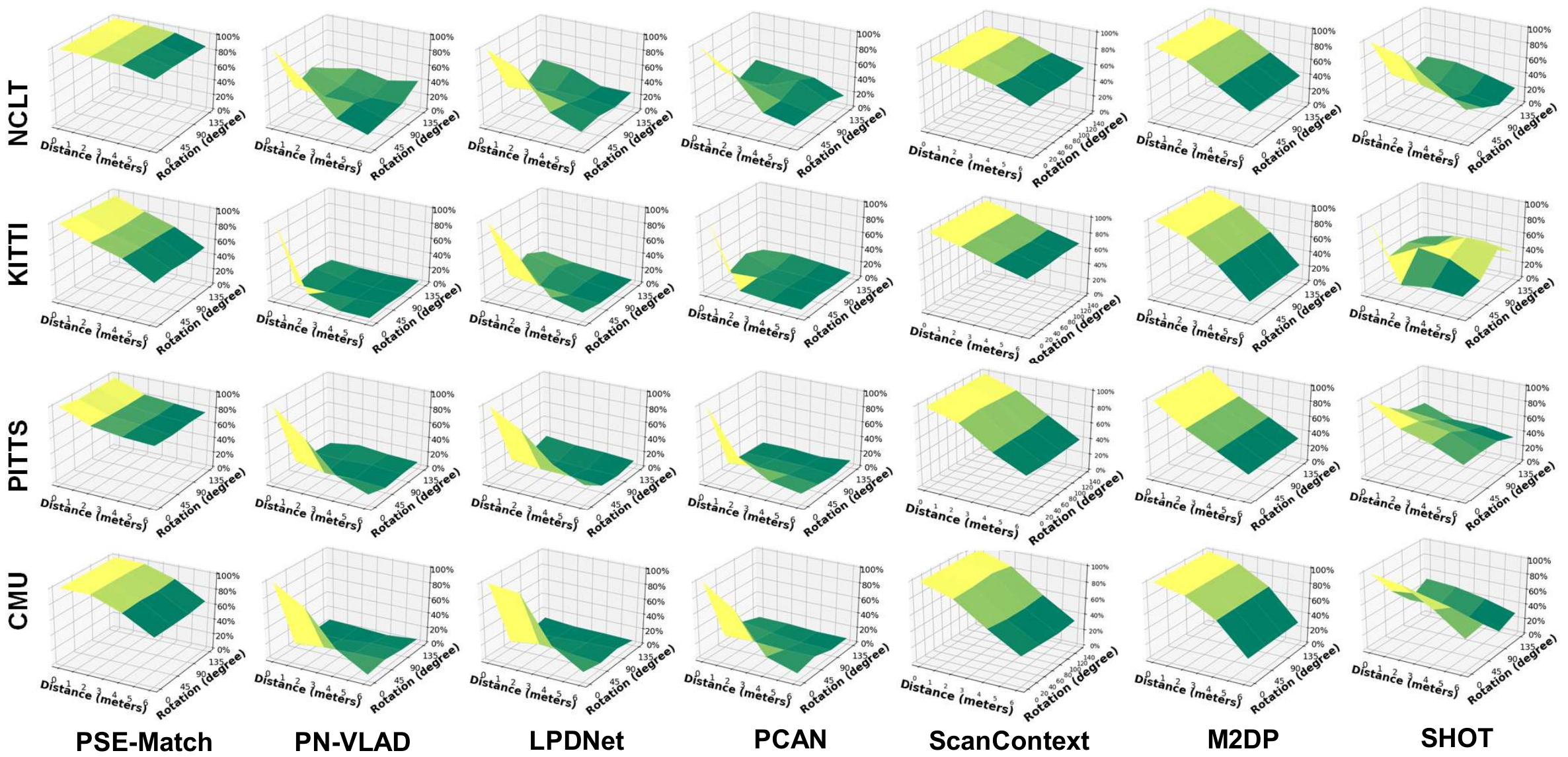}
        \caption{Average recall at top $1\%$ under variant viewpoint differences.
        For each sub-graph, x-axis represents the translation differences between the testing and reference trajectories, y-axis represents the relative orientation differences. z-axis shows the relative average recall.}
        \label{fig:recall}
    \end{figure*}
    
    We evaluate the place recognition accuracy under two different matching methods: single-scan and sequential-scan based place recognition.
    The first method retrieves corresponding places by evaluating feature similarity with K-nearest neighbor search.
    The second method defines a temporal window to capture more stable place features, and use brute-force searching to locate the best matches~\cite{VPR:SeqSLAM}.
    
    \bigskip
    \noindent \textbf{Evaluation Metrics.}
    In both evaluation methods, similar to PN-VLAD~\cite{PR:pointnetvlad}, we use the Average Recall at top one, ten and $1\%$ retrievals, the Precision-Recall Curve and Average Precision Score (AP score) under different viewpoints to assess the place recognition accuracy between the relative reference and testing sequence trajectories.

\subsection{Place Recognition Results}
\label{sec:PR_results}

\subsubsection{Orientation- and Translation- Invariant Analysis}

    Fig.~\ref{fig:dataset} demonstrate the top one place retrievals of different methods with a given reference map, we can see that PSE-Match can retrieve the correct sub-maps under both orientation and translation differences.
    Fig.~\ref{fig:recall} gives a more detailed recognition performance, where each sub-figure represents the place recognition analysis with a fixed viewpoint difference.
    The x-axis represents the translation difference (from $0\sim 10$m), y-axis represents the orientation difference (from $0\sim180^{\circ}$), and z-axis represents the average recall at top $1\%$ retrieves.
    For all the non-learning based approaches, we can note that: these methods are robust to orientation differences, but weak to locally translation differences.
    Among the non-learning methods, both ScanContext and M2DP show robuster place recognition ability under variant orientation differences; however, this method is sensitive to the locally translation differences in the XY-plane.
    This performance of the non-learning based methods is slightly different from the other work~\cite{locnet,scan_context}, because the viewpoint didn't includes variant orientation differences.
    Among all the other non-learning methods, ScanContext shows the best performance, but it has a higher time-consumption as we analyzed in Section.~\ref{sec:exp_runtime}.

    \begin{figure*}[t]
        \centering
        \subfloat[Single-scan]
        {{\includegraphics[width=0.48\linewidth]{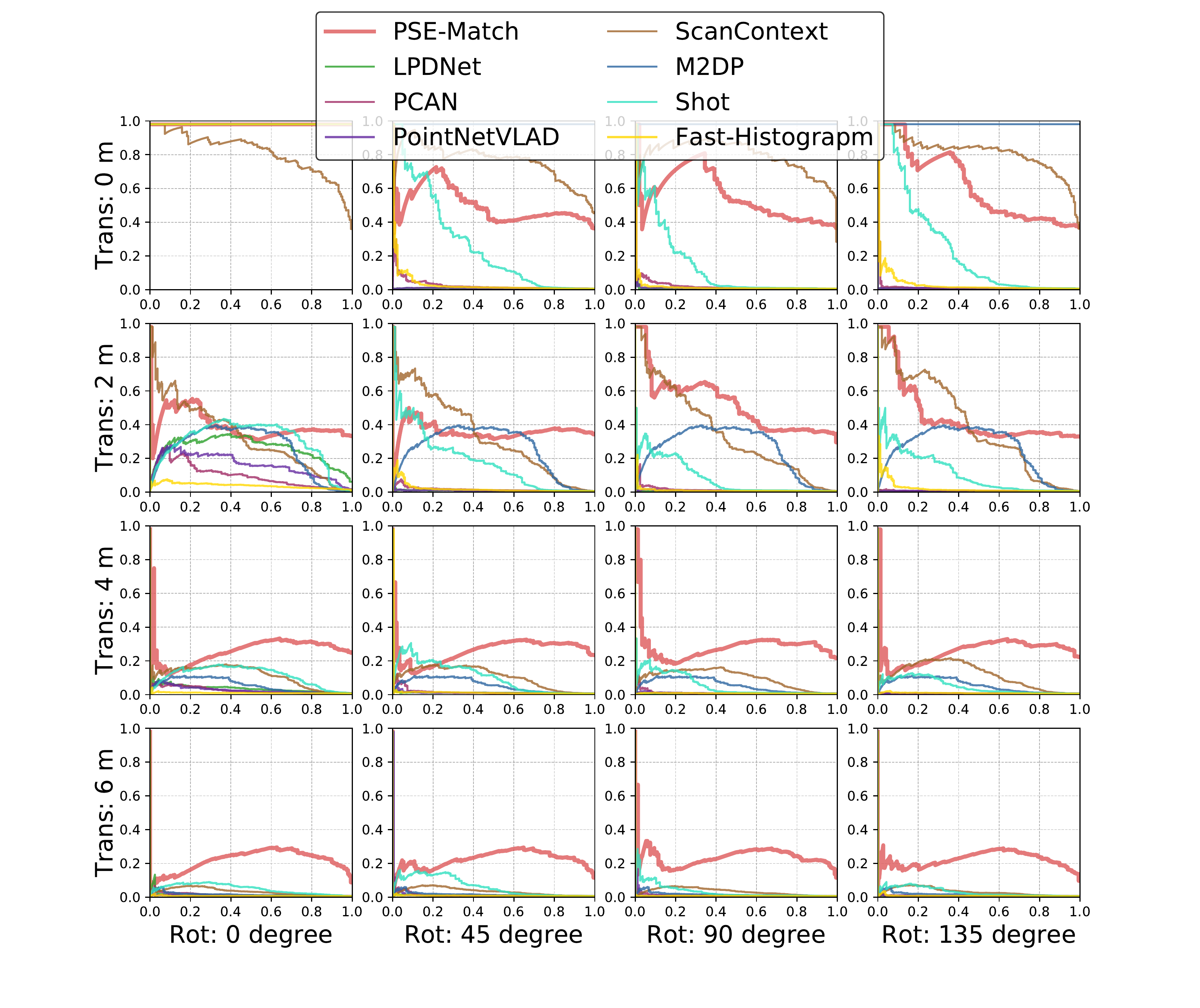} }}%
        \subfloat[Sequential-scan]
        {{\includegraphics[width=0.48\linewidth]{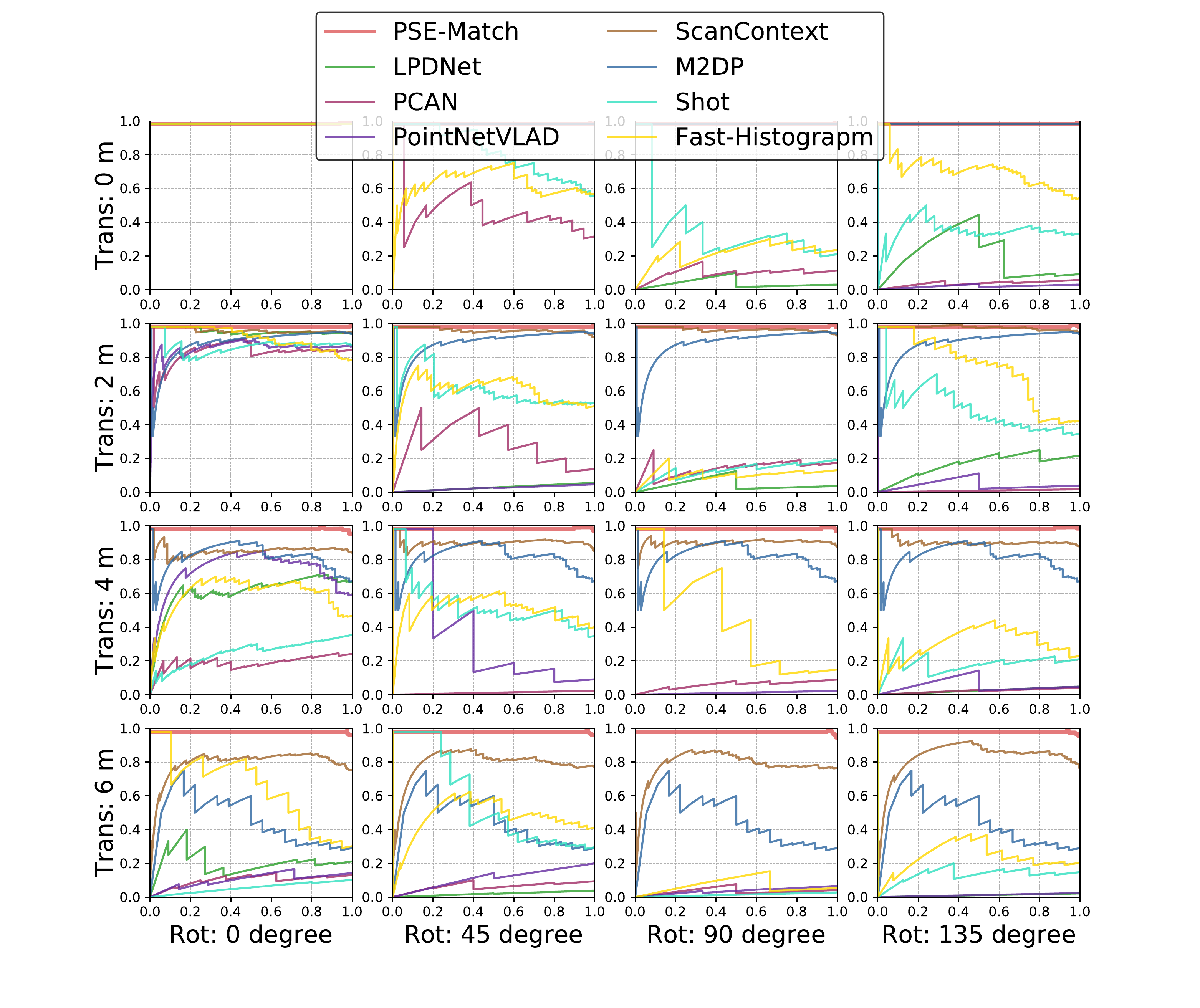} }}%
        \caption{Precision-recall curves of single/sequential-scan matching on the \textit{NCLT} dataset.}
        \label{fig:pr_curve}%
    \end{figure*}

    \begin{figure*}[ht]
        \centering
        \subfloat[Single-scan]
        {{\includegraphics[width=0.48\linewidth]{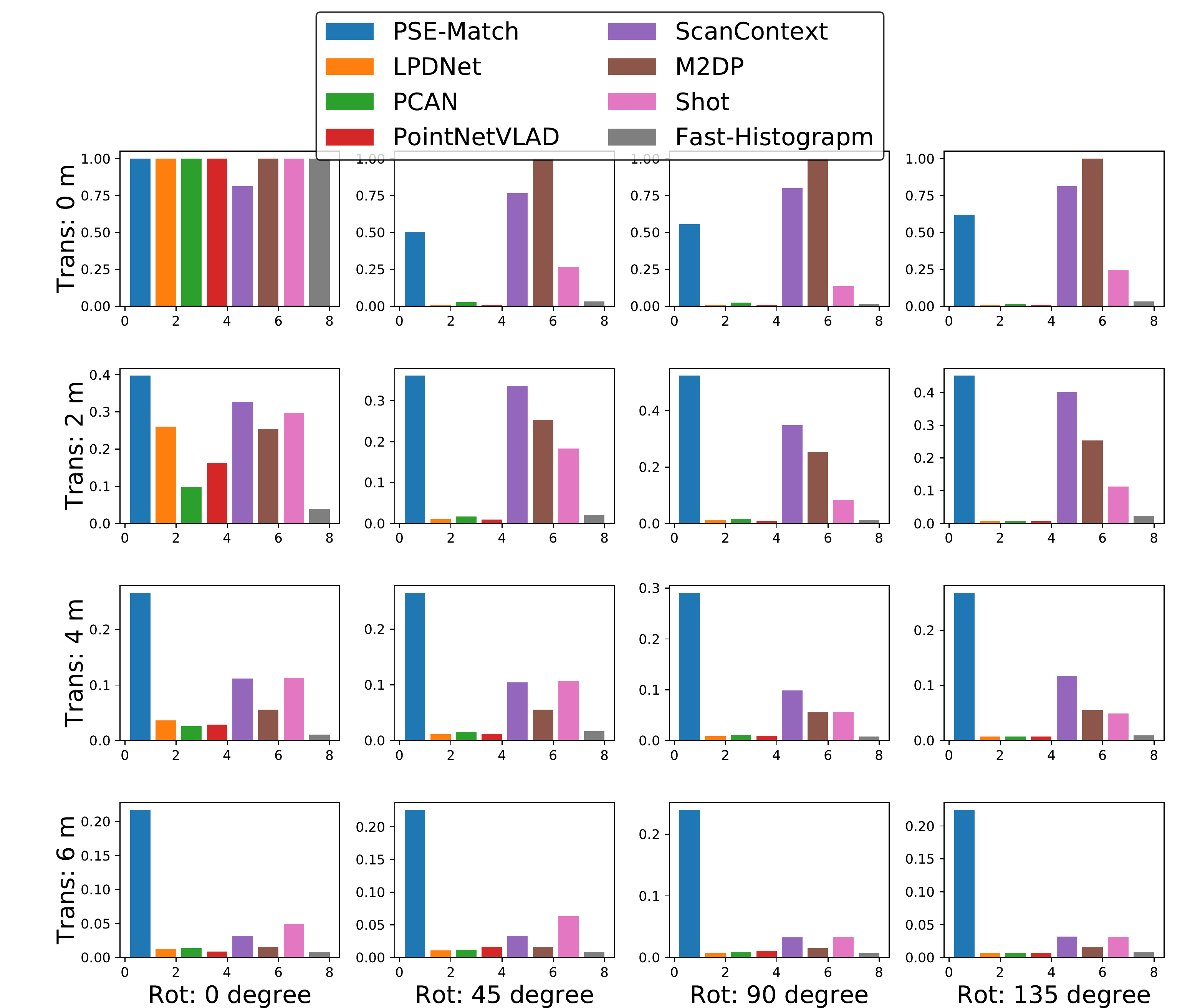} }}%
        \subfloat[Sequential-scan]
        {{\includegraphics[width=0.48\linewidth]{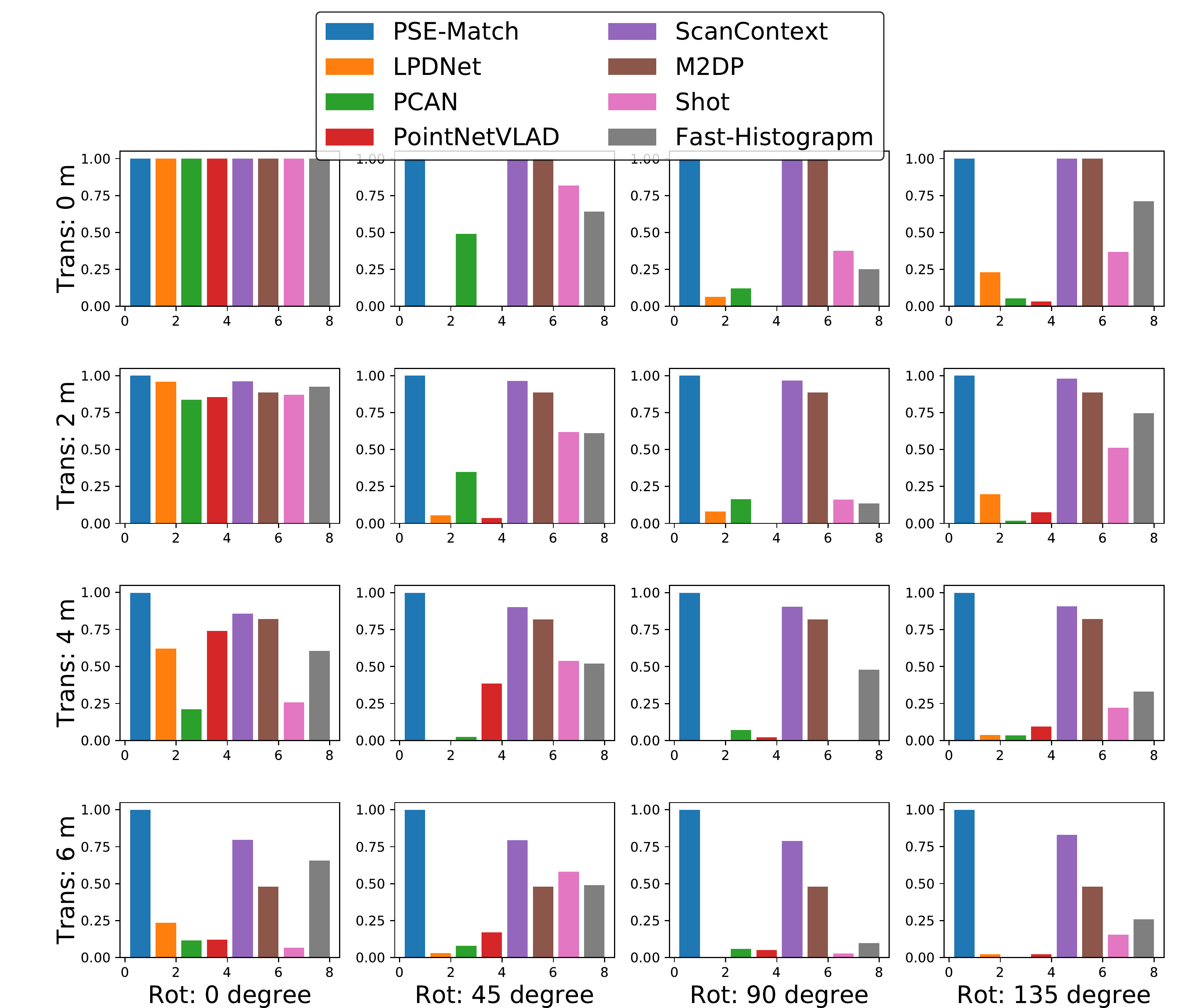} }}%
        \caption{Average precision score (AP score) of single/sequential-scan matching on the \textit{NCLT} dataset.}
    \label{fig:auc_score}%
    \end{figure*}

  \begin{table}[ht]
        \centering
        \caption{The average recall of at top one on four datasets with translation differences range from $0\sim8$m and orientation changes range from $0\sim135^{\circ}$.}
        \begin{tabular}{c c c c c c c}
        \toprule
        Method & \textit{NCLT} & \textit{KITTI} & \textit{City} & \textit{CMU} \\ \midrule
        ScanContext~\cite{scan_context} & $63.19\%$ & $61.27\%$ & $60.81\%$ & $64.19\%$ \\
        M2DP~\cite{he2016m2dp}          & $59.23\%$ & $58.99\%$ & $57.03\%$ & $64.32\%$ \\
        SHOT~\cite{shot}                & $33.33\%$ & $30.33\%$ & $48.98\%$ & $44.47\%$ \\
        Fast-Histogram~\cite{histogram} & $14.23\%$ & $18.83\%$ & $12.32\%$ & $15.19\%$ \\
        \hline
        PN-VLAD~\cite{PR:pointnetvlad}             & $13.85\%$ & $7.39\%$ & $14.97\%$ & $13.98\%$ \\ 
        LPDNet~\cite{PR:LPDNet}              & $13.10\%$ & $12.92\%$ & $18.63\%$ & $17.26\%$ \\ 
        PCAN~\cite{PR:PCAN}                & $13.02\%$ & $9.80\%$ & $13.02\%$ & $15.47\%$ \\ 
        PSE-Match (no DOM)      & $67.21\%$  & $64.98\%$  & $70.71\%$ & $67.13\%$ \\
        PSE-Match               & $\mathbf{82.78\%}$ & $\mathbf{72.19\%}$ & $\mathbf{78.06\%}$ & $\mathbf{75.95\%}$ \\
        \hline
        PN-VLAD (refine)    & $12.67\%$ & $7.33\%$ & $13.54\%$ & $14.97\%$ \\ 
        LPDNet (refine)     & $12.07\%$ & $8.18\%$ & $12.01\%$ & $13.41\%$ \\
        PCAN (refine)       & $11.89\%$ & $9.89\%$ & $14.25\%$ & $16.08\%$ \\
        PSE-Match (refine)  & $72.33\%$ & $69.12\%$ & $70.91\%$ & $70.61\%$ \\
        \bottomrule
        \label{table:ar_1p}
        \end{tabular}
    \end{table}

    On the other hand, for the current learning-based place recognition methods, they shows a better robustness to locally translation differences than the non-learning ones.
    However, as we can see in Fig.~\ref{fig:recall}, all PointNet-based place features are not aimed at dealing with orientation-invariance, thus they are also very sensitive to the orientation differences.
    Table.~\ref{table:ar_1p} and Table.~\ref{table:ar_10p} give the detailed average recall at top one and ten respectively, and please note the recalls are averaged all the viewpoint differences (i.e., translation differences range from $0\sim8$m and orientation changes range from $0\sim135^{\circ}$).
    As we can note that, both LPDNet~\cite{PR:LPDNet} and PCAN~\cite{PR:PCAN} show better place recognition ability than the original PN-VLAD~\cite{PR:pointnetvlad} under orientation differences.
    However, they cannot handle significant perspective changes in both translation and orientation simultaneously.

    \begin{table}[t]
        \centering
        \caption{The average recall of at top ten on four datasets with translation differences range from $0\sim8$m and orientation changes range from $0\sim135^{\circ}$.}
        \begin{tabular}{c c c c c c c}
        \toprule
        Method & \textit{KITTI} & \textit{NCLT} & \textit{PITTS} & \textit{CMU} \\ \midrule
        ScanContext~\cite{scan_context}  & $85.21\%$ & $90.15\%$ & $85.88\%$ & $88.49\%$ \\
        M2DP~\cite{he2016m2dp}  & $83.39\%$ & $83.53\%$ & $80.68\%$ & $85.82\%$ \\
        SHOT~\cite{shot}        & $55.87\%$ & $53.26\%$ & $74.73\%$ & $68.55\%$ \\
        Fast-Histogram~\cite{histogram} & $31.31\%$ & $21.43\%$ & $28.07\%$ & $30.98\%$ 
        \\ 
        \hline
        PN-VLAD~\cite{PR:pointnetvlad}  & $36.42\%$  & $13.21\%$  & $33.90\%$ & $32.24\%$ \\ 
        LPDNet~\cite{PR:LPDNet}         & $34.18\%$  & $29.05\%$  & $37.05\%$ & $35.13\%$ \\ 
        PCAN~\cite{PR:PCAN}             & $38.85\%$  & $23.36\%$  & $32.70\%$ & $38.45\%$ \\ 
        PSE-Match (no DOM)      & $88.32\%$  & $89.09\%$  & $93.21\%$ & $91.31\%$ \\
        PSE-Match       & $\mathbf{99.89\%}$ & $\mathbf{96.28\%}$ & $\mathbf{99.16\%}$ & $\mathbf{99.50\%}$ \\
        \hline
        PN-VLAD (refine)    & $30.77\%$ & $10.70\%$ & $27.73\%$ & $29.15\%$ \\ 
        LPDNet (refine)     & $30.09\%$ & $14.98\%$ & $23.42\%$ & $28.20\%$ \\
        PCAN (refine)       & $34.47\%$ & $21.13\%$ & $34.69\%$ & $38.45\%$ \\
        PSE-Match (refine)  & $97.16\%$ & $94.72\%$ & $96.40\%$ & $97.17\%$ \\
        \bottomrule
        \label{table:ar_10p}
        \end{tabular}
    \end{table}
    
    In contrast, PSE-Match shows better performance in place recognition under variant viewpoint differences.
    On all the four datasets, the average recalls at top one are above $70\%$; and in the refine mode, PSE-Match also shows better generalization ability than both non-learning/learning based methods.
    In the top ten retrieval case, PSE-Match can reach above $95\%$ on all datasets.
    We also analysis the place recognition performance of our dynamic octree mapping (DOM) module. 
    The PSE-Match (no DOM) in Table.~\ref{table:ar_1p} and Table.~\ref{table:ar_10p} indicate the local mapping using accumulated point clouds, without the measurement and motion updating in the DOM module.
    In both the recalls of top one and ten retrievals, PSE-Match can provide more reliable place recognition with DOM module.
    
    \begin{figure*}[t]
    	\centering
    	\includegraphics[width=\linewidth]{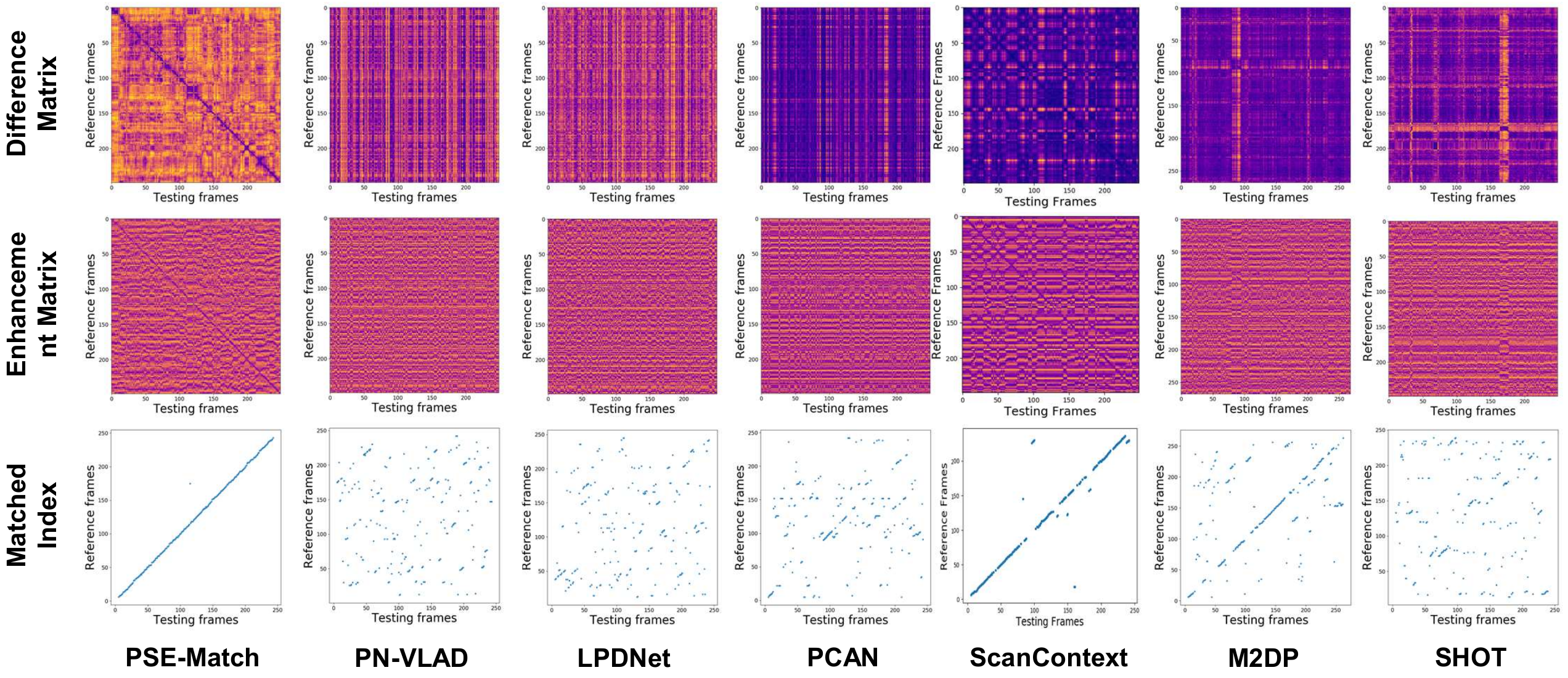}
    	\caption{Sequence matching performance of different methods on \textit{KITTI} datasets.
    	The first row shows the cosine feature difference between testing and reference trajectory, and the relative translation and orientation difference between two trajectories is $4m$ and $135^{\circ}$ respectively.
    	The second row shows the enhanced difference matrix by apply local normalization same to the original SeqSLAM~\cite{VPR:SeqSLAM}.
    	The last row shows the estimated sequential matching results.}
    	\label{fig:seqslam}
    \end{figure*}
    
    \subsubsection{Single and Sequential Matching Results}
    In this subsection, we further investigate the place recognition performance under single-scan and sequential-scan matching mode.
    Fig.~\ref{fig:pr_curve} and Fig.~\ref{fig:auc_score} shows the precision-recall curve and average precision score of different methods on the \textit{NCLT} dataset.
    In the cases of lower translation and orientation cases, all methods can achieve robust place recognition ability.
    And both ScanContext and M2DP show the invariant property to orientation and perform well under smaller translation and orientation.
    However, for significant translational or rotational cases, both transitional methods or PointNet-based method cannot guarantee accurate place retrieval.
    Especially, the PointNet based methods are very sensitive to the orientation differences.

    Obviously, the sequence-matching manner can improve the matching performances.
    The performance of both ScanContext and M2DP are significantly improved.
    However, PointNet-based methods still fail to find the matches in the cases with big orientation differences.
    For PSE-Match method, it can achieve even better recognition performance with the sequential matching manner.
    Fig.~\ref{fig:seqslam} shows the feature difference matrix, enhancement matrix and matched indexes in the sequential-matching~\cite{VPR:SeqSLAM} procedure.
    We can see the reference/testing features difference matrix has a stronger contrast than other methods, which provides stable sequential matches.

\subsection{Time and Storage Analysis}
    \label{sec:exp_runtime}
    
    In Table.~\ref{table:time_efficiency}, we analysis the time and memory usage comparison of our PSE-Match and other non-learning/learning based place recognition methods.
    Both ScanContext and M2DP is time consuming in feature extraction.
    Compared with ScanContext, M2DP and other learning-based methods, PSE-Match consumes less GPU memory in the training procedure.
    In the inference step, we running the dynamic octree mapping module to generate dense local map, which takes $10$ms in application.
    The semantic inference and PSE-Match feature extraction for each sub-map will take $15ms$ ($250MB$ GPU) and $13.24ms$ ($37Mb$ GPU) respectively.
    {It is feasible to run PSE-Match in the Nvidia embedded system (Nvidia Jetson TX2, or Xavier), and suitable for low-cost robot systems in long-term SLAM and navigation.
    }
    
    \begin{table}[t]
        \centering
        \caption{Comparison result of time and memory requirements of different methods.}
        \begin{tabular}{c c c c c}
        \toprule
        Method      & GPU (M)  & Time (ms) & Feature Size  \\ \midrule
        \textit{ScanContext}  &  $-$   & $258.29$    & 192    \\ 
        \textit{M2DP}  &  $-$   & $164.05$    & 192    \\ 
        \textit{SHOT}  &  $-$   & $14.74$     & 352    \\ 
        \textit{Fast Histogram}  &  $-$   & $2.53$     & 192    \\ 
        \hline
        \textit{PN-VLAD~\cite{PR:pointnetvlad}}  &  $2887$    & $11.31$ & 256 \\ 
        \textit{LPDNet~\cite{PR:LPDNet}}  &  $4679$    & $41.23$ & 256 \\ 
        \textit{PCAN~\cite{PR:PCAN}}  &  $8525$    & $21.53$ & 256 \\ 
        \textit{PSE-Match}    &  $2459$    & $10+15+13.24$ & 256 \\ 
        \bottomrule 
        \end{tabular}
        \label{table:time_efficiency}
    \end{table}
    
\section{Conclusions}
\label{sec:conclusions}
    In this paper, we propose PSE-Match, a 3D place recognition and global localization method.
    Compared with the current 3D place recognition methods, PSE-Match can learn robust and viewpoint-free 3D feature descriptors from different independent semantic attributes, which can provide stable consistence in complex 3D environment and avoid the disturbance from the dynamic objects.
    Particularly, we introduce a parallel semantic embedding network to learn viewpoint-free feature descriptor on the spherical projections with assistance of a divergence learning metric. 
    Validating on both single and sequential matching data, experiments results show that our method has significant accuracy improvement than other state-of-the-arts peers under variant viewpoint differences.
    The less computation resource requirements and efficient inference make our method more practical for real-time applications.
    
    In the future work, we will extend the large-scale loop closure detection and mapping based on the PSE-Match.
    Leveraging the native properties of semantic embedding, we will pursue embedding objective-level information for accurate localization for autonomous navigation.

\color{black}
\bibliographystyle{IEEEtranTIE}
\bibliography{Refer}\ 

\begin{thebibliography}{10}
\providecommand{\url}[1]{#1}
\csname url@samestyle\endcsname
\providecommand{\newblock}{\relax}
\providecommand{\bibinfo}[2]{#2}
\providecommand{\BIBentrySTDinterwordspacing}{\spaceskip=0pt\relax}
\providecommand{\BIBentryALTinterwordstretchfactor}{4}
\providecommand{\BIBentryALTinterwordspacing}{\spaceskip=\fontdimen2\font plus
\BIBentryALTinterwordstretchfactor\fontdimen3\font minus
  \fontdimen4\font\relax}
\providecommand{\BIBforeignlanguage}[2]{{%
\expandafter\ifx\csname l@#1\endcsname\relax
\typeout{** WARNING: IEEEtran.bst: No hyphenation pattern has been}%
\typeout{** loaded for the language `#1'. Using the pattern for}%
\typeout{** the default language instead.}%
\else
\language=\csname l@#1\endcsname
\fi
#2}}
\providecommand{\BIBdecl}{\relax}
\BIBdecl

\bibitem{Survey:VSLAM}
J.~Fuentes-Pacheco, J.~Ruiz-Ascencio, and J.~M. Rend{\'o}n-Mancha, ``Visual
  simultaneous localization and mapping: a survey,'' \emph{Artificial
  intelligence review}, vol.~43, no.~1, pp. 55--81, 2015.

\bibitem{Auto:larson2019autonomous}
E.~Larson, A.~Ianuzzi, R.~Oliva, D.~Lowrance \emph{et~al.}, ``Autonomous
  underwater survey apparatus and system,'' May~2 2019, uS Patent App.
  16/173,567.

\bibitem{Auto:perception}
\BIBentryALTinterwordspacing
F.~Rosique, P.~J. Navarro, C.~Fernández, and A.~Padilla, ``A systematic review
  of perception system and simulators for autonomous vehicles research,''
  \emph{Sensors}, vol.~19, no.~3, 2019. [Online]. Available:
  \url{https://www.mdpi.com/1424-8220/19/3/648}
\BIBentrySTDinterwordspacing

\bibitem{Auto:control}
W.~Lim, S.~Lee, M.~Sunwoo, and K.~Jo, ``Hierarchical trajectory planning of an
  autonomous car based on the integration of a sampling and an optimization
  method,'' \emph{IEEE Transactions on Intelligent Transportation Systems},
  vol.~19, \href{http://dx.doi.org/10.1109/TITS.2017.2756099}{DOI
  10.1109/TITS.2017.2756099}, no.~2, pp. 613--626, 2018.

\bibitem{VPR:FABMAP}
M.~{Nowakowski}, C.~{Joly}, S.~{Dalibard}, N.~{Garcia}, and F.~{Moutarde},
  ``Topological localization using wi-fi and vision merged into fabmap
  framework,'' in \emph{2017 IEEE/RSJ International Conference on Intelligent
  Robots and Systems (IROS)}, pp. 3339--3344, 2017.

\bibitem{VPR:ORB-SLAM}
W.~Zong, L.~Chen, C.~Zhang, Z.~Wang, and Q.~Chen, ``{Vehicle model based
  visual-tag monocular ORB-SLAM},'' in \emph{IEEE International Conference on
  Systems, Man, and Cybernetics (SMC)},
  \href{http://dx.doi.org/10.1109/SMC.2017.8122816}{DOI
  10.1109/SMC.2017.8122816}, pp. 1441--1446, Oct. 2017.

\bibitem{FeatureCapturer:BoW2}
H.~{Jégou}, F.~{Perronnin}, M.~{Douze}, J.~{Sánchez}, P.~{Pérez}, and
  C.~{Schmid}, ``Aggregating local image descriptors into compact codes,''
  \emph{IEEE Transactions on Pattern Analysis and Machine Intelligence},
  vol.~34, no.~9, pp. 1704--1716, 2012.

\bibitem{FEATURE:ORB}
E.~Rublee, V.~Rabaud, K.~Konolige, and G.~Bradski, ``{ORB: An efficient
  alternative to SIFT or SURF},'' in \emph{International Conference on Computer
  Vision}, \href{http://dx.doi.org/10.1109/ICCV.2011.6126544}{DOI
  10.1109/ICCV.2011.6126544}, pp. 2564--2571, Nov. 2011.

\bibitem{FEATURE:SURF}
Z.~Zhang, Y.~Huang, C.~Li, and Y.~Kang, ``{Monocular vision simultaneous
  localization and mapping using SURF},'' in \emph{7th World Congress on
  Intelligent Control and Automation},
  \href{http://dx.doi.org/10.1109/WCICA.2008.4593166}{DOI
  10.1109/WCICA.2008.4593166}, pp. 1651--1656, Jun. 2008.

\bibitem{PR:netvlad}
R.~{Arandjelovic}, P.~{Gronat}, A.~{Torii}, T.~{Pajdla}, and J.~{Sivic},
  ``Netvlad: Cnn architecture for weakly supervised place recognition,'' in
  \emph{2016 IEEE Conference on Computer Vision and Pattern Recognition
  (CVPR)}, \href{http://dx.doi.org/10.1109/CVPR.2016.572}{DOI
  10.1109/CVPR.2016.572}, pp. 5297--5307, Jun. 2016.

\bibitem{VPR:SeqSLAM}
M.~J. Milford and G.~F. Wyeth, ``{SeqSLAM: Visual route-based navigation for
  sunny summer days and stormy winter nights},'' in \emph{IEEE International
  Conference on Robotics and Automation},
  \href{http://dx.doi.org/10.1109/ICRA.2012.6224623}{DOI
  10.1109/ICRA.2012.6224623}, pp. 1643--1649, May. 2012.

\bibitem{segmap}
R.~Dubé, A.~Cramariuc, D.~Dugas, H.~Sommer, M.~Dymczyk, J.~Nieto, R.~Siegwart,
  and C.~Cadena, ``Segmap: Segment-based mapping and localization using
  data-driven descriptors,'' \emph{The International Journal of Robotics
  Research}, vol.~39, \href{http://dx.doi.org/10.1177/0278364919863090}{DOI
  10.1177/0278364919863090}, no. 2-3, pp. 339--355, 2020.

\bibitem{PointNet}
C.~R. Qi, H.~Su, K.~Mo, and L.~J. Guibas, ``Pointnet: Deep learning on point
  sets for 3d classification and segmentation,'' in \emph{Proceedings of the
  IEEE Conference on Computer Vision and Pattern Recognition}, pp. 652--660,
  2017.

\bibitem{pointnet++}
C.~R. Qi, L.~Yi, H.~Su, and L.~J. Guibas, ``Pointnet++: Deep hierarchical
  feature learning on point sets in a metric space,'' in \emph{Advances in
  neural information processing systems}, pp. 5099--5108, 2017.

\bibitem{PR:pointnetvlad}
M.~Angelina~Uy and G.~Hee~Lee, ``Pointnetvlad: Deep point cloud based retrieval
  for large-scale place recognition,'' in \emph{The IEEE Conference on Computer
  Vision and Pattern Recognition (CVPR)}, Jun. 2018.

\bibitem{PR:PCAN}
W.~{Zhang} and C.~{Xiao}, ``Pcan: 3d attention map learning using contextual
  information for point cloud based retrieval,'' in \emph{2019 IEEE/CVF
  Conference on Computer Vision and Pattern Recognition (CVPR)}, pp.
  12\,428--12\,437, 2019.

\bibitem{PR:LPDNet}
Z.~Liu, S.~Zhou, C.~Suo, P.~Yin, W.~Chen, H.~Wang, H.~Li, and Y.-H. Liu,
  ``Lpd-net: 3d point cloud learning for large-scale place recognition and
  environment analysis,'' in \emph{Proceedings of the IEEE International
  Conference on Computer Vision}, pp. 2831--2840, 2019.

\bibitem{lowe1999object}
D.~G. Lowe, ``Object recognition from local scale-invariant features,'' in
  \emph{Proceedings of the seventh IEEE international conference on computer
  vision}, vol.~2, pp. 1150--1157.\hskip 1em plus 0.5em minus 0.4em\relax Ieee,
  1999.

\bibitem{bay2006surf}
H.~Bay, T.~Tuytelaars, and L.~Van~Gool, ``Surf: Speeded up robust features,''
  in \emph{Computer Vision -- ECCV 2006}, A.~Leonardis, H.~Bischof, and
  A.~Pinz, Eds., pp. 404--417.\hskip 1em plus 0.5em minus 0.4em\relax Berlin,
  Heidelberg: Springer Berlin Heidelberg, 2006.

\bibitem{AG:SpinImageNew}
Y.~Mei and Y.~He, ``A new spin-image based 3d map registration algorithm using
  low-dimensional feature space,'' in \emph{IEEE International Conference on
  Information and Automation (ICIA)},
  \href{http://dx.doi.org/10.1109/ICInfA.2013.6720358}{DOI
  10.1109/ICInfA.2013.6720358}, pp. 545--551, Aug. 2013.

\bibitem{esf}
W.~Wohlkinger and M.~Vincze, ``Ensemble of shape functions for 3d object
  classification,'' in \emph{2011 IEEE international conference on robotics and
  biomimetics}, pp. 2987--2992.\hskip 1em plus 0.5em minus 0.4em\relax IEEE,
  2011.

\bibitem{scan_context}
G.~Kim and A.~Kim, ``Scan context: Egocentric spatial descriptor for place
  recognition within 3d point cloud map,'' in \emph{2018 IEEE/RSJ International
  Conference on Intelligent Robots and Systems (IROS)}, pp. 4802--4809.\hskip
  1em plus 0.5em minus 0.4em\relax IEEE, 2018.

\bibitem{shot}
F.~Tombari, S.~Salti, and L.~Di~Stefano, ``Unique signatures of histograms for
  local surface description,'' in \emph{Computer Vision -- ECCV 2010},
  K.~Daniilidis, P.~Maragos, and N.~Paragios, Eds., pp. 356--369.\hskip 1em
  plus 0.5em minus 0.4em\relax Berlin, Heidelberg: Springer Berlin Heidelberg,
  2010.

\bibitem{rohling2015fast}
T.~R{\"o}hling, J.~Mack, and D.~Schulz, ``A fast histogram-based similarity
  measure for detecting loop closures in 3-d lidar data,'' in \emph{2015
  IEEE/RSJ International Conference on Intelligent Robots and Systems (IROS)},
  pp. 736--741.\hskip 1em plus 0.5em minus 0.4em\relax IEEE, 2015.

\bibitem{he2016m2dp}
L.~He, X.~Wang, and H.~Zhang, ``M2dp: A novel 3d point cloud descriptor and its
  application in loop closure detection,'' in \emph{2016 IEEE/RSJ International
  Conference on Intelligent Robots and Systems (IROS)}, pp. 231--237.\hskip 1em
  plus 0.5em minus 0.4em\relax IEEE, 2016.

\bibitem{locnet}
H.~Yin, Y.~Wang, X.~Ding, L.~Tang, S.~Huang, and R.~Xiong, ``3d lidar-based
  global localization using siamese neural network,'' \emph{IEEE Transactions
  on Intelligent Transportation Systems}, vol.~21, no.~4, pp. 1380--1392, 2019.

\bibitem{spherecnn}
C.~Esteves, C.~Allen-Blanchette, A.~Makadia, and K.~Daniilidis, ``Learning so
  (3) equivariant representations with spherical cnns,'' in \emph{Proceedings
  of the European Conference on Computer Vision (ECCV)}, pp. 52--68, 2018.

\bibitem{maskrcnn}
K.~He, G.~Gkioxari, P.~Doll{\'a}r, and R.~Girshick, ``Mask r-cnn,'' in
  \emph{Proceedings of the IEEE international conference on computer vision},
  pp. 2961--2969, 2017.

\bibitem{mink}
C.~Choy, J.~Gwak, and S.~Savarese, ``4d spatio-temporal convnets: Minkowski
  convolutional neural networks,'' in \emph{Proceedings of the IEEE Conference
  on Computer Vision and Pattern Recognition}, pp. 3075--3084, 2019.

\bibitem{schonberger2018semantic}
J.~L. Sch{\"o}nberger, M.~Pollefeys, A.~Geiger, and T.~Sattler, ``Semantic
  visual localization,'' in \emph{Proceedings of the IEEE Conference on
  Computer Vision and Pattern Recognition}, pp. 6896--6906, 2018.

\bibitem{semantic-geo}
\BIBentryALTinterwordspacing
S.~Garg, N.~Suenderhauf, and M.~Milford, ``Semantic geometric visual place
  recognition: a new perspective for reconciling opposing views,'' \emph{The
  International Journal of Robotics Research}, vol.~0,
  \href{http://dx.doi.org/10.1177/0278364919839761}{DOI
  10.1177/0278364919839761}, no.~0, p. 0278364919839761, 0. [Online].
  Available: \url{https://doi.org/10.1177/0278364919839761}
\BIBentrySTDinterwordspacing

\bibitem{3DSegL:wu2018squeezeseg}
B.~{Wu}, A.~{Wan}, X.~{Yue}, and K.~{Keutzer}, ``Squeezeseg: Convolutional
  neural nets with recurrent crf for real-time road-object segmentation from 3d
  lidar point cloud,'' in \emph{2018 IEEE International Conference on Robotics
  and Automation (ICRA)},
  \href{http://dx.doi.org/10.1109/ICRA.2018.8462926}{DOI
  10.1109/ICRA.2018.8462926}, pp. 1887--1893, May. 2018.

\bibitem{LOAM:Lego-LOAM}
T.~{Shan} and B.~{Englot}, ``Lego-loam: Lightweight and ground-optimized lidar
  odometry and mapping on variable terrain,'' in \emph{2018 IEEE/RSJ
  International Conference on Intelligent Robots and Systems (IROS)},
  \href{http://dx.doi.org/10.1109/IROS.2018.8594299}{DOI
  10.1109/IROS.2018.8594299}, pp. 4758--4765, Oct. 2018.

\bibitem{LPR:OCTOMAP}
E.~Rho and S.~Jo, ``Octomap-based semi-autonomous quadcopter navigation with
  biosignal classification,'' in \emph{6th International Conference on
  Brain-Computer Interface (BCI)},
  \href{http://dx.doi.org/10.1109/IWW-BCI.2018.8311533}{DOI
  10.1109/IWW-BCI.2018.8311533}, pp. 1--4, Jan. 2018.

\bibitem{ROBOT:probrobo}
S.~Thrun, ``Probabilistic robotics,'' \emph{Communications of the ACM},
  vol.~45, no.~3, pp. 52--57, 2002.

\bibitem{LOAM:zhang2014loam}
J.~Zhang and S.~Singh, ``Loam: Lidar odometry and mapping in real-time.'' in
  \emph{Robotics: Science and Systems}, vol.~2, p.~9, 2014.

\bibitem{Sphere:SO3_invariant}
M.~Kazhdan, T.~Funkhouser, and S.~Rusinkiewicz, ``Rotation invariant spherical
  harmonic representation of 3 d shape descriptors,'' in \emph{Symposium on
  geometry processing}, vol.~6, pp. 156--164, 2003.

\bibitem{cohen2018spherical}
T.~S. Cohen, M.~Geiger, J.~K{\"o}hler, and M.~Welling, ``Spherical cnns,'' in
  \emph{International Conference on Learning Representations}, 2018.

\bibitem{PR:vlad}
R.~{Arandjelovic} and A.~{Zisserman}, ``All about vlad,'' in \emph{2013 IEEE
  Conference on Computer Vision and Pattern Recognition},
  \href{http://dx.doi.org/10.1109/CVPR.2013.207}{DOI 10.1109/CVPR.2013.207},
  pp. 1578--1585, Jun. 2013.

\bibitem{KITTI:Geiger}
A.~Geiger, P.~Lenz, C.~Stiller, and R.~Urtasun, ``{Vision meets robotics: The
  KITTI dataset},'' \emph{{INTERNATIONAL JOURNAL OF ROBOTICS RESEARCH}},
  vol.~{32}, \href{http://dx.doi.org/{10.1177/0278364913491297}}{DOI
  {10.1177/0278364913491297}}, no.~{11}, pp. {1231--1237}, {SEP} {2013}.

\bibitem{DATASET:NCTL}
N.~Carlevaris-Bianco, A.~K. Ushani, and R.~M. Eustice, ``University of michigan
  north campus long-term vision and lidar dataset,'' \emph{The International
  Journal of Robotics Research}, vol.~35, no.~9, pp. 1023--1035, 2016.

\bibitem{histogram}
R.~B. Rusu, N.~Blodow, and M.~Beetz, ``Fast point feature histograms (fpfh) for
  3d registration,'' in \emph{2009 IEEE International Conference on Robotics
  and Automation}, pp. 3212--3217.\hskip 1em plus 0.5em minus 0.4em\relax IEEE,
  2009.

\end{thebibliography}

\end{document}